%% file: main.tex
\documentclass[10pt]{article} %

\usepackage{amsthm}
\usepackage{newpxtext}
\usepackage{geometry} %
\usepackage{natbib}
\PassOptionsToPackage{numbers, sort&compress}{natbib}

\usepackage[utf8]{inputenc} %
\usepackage[T1]{fontenc}    %
\usepackage{hyperref}       %
\usepackage{url}            %
\usepackage{booktabs}       %
\usepackage{amsfonts}       %
\usepackage{microtype}      %

\usepackage{abstract}

\usepackage{xcolor}

\definecolor{mydarkblue}{rgb}{0,0.08,0.45}
\definecolor{mydarkgreen}{rgb}{0,0.45,0.08}
  \hypersetup{ %
    colorlinks=true,
    linkcolor=mydarkblue,
    citecolor=mydarkblue,
    filecolor=mydarkblue,
    urlcolor=mydarkblue}

\input{math_commands.tex}

\usepackage{wrapfig}

\usepackage{float}

\usepackage{amsthm}
\usepackage{amssymb}

\usepackage{enumitem}

\usepackage{graphicx}

\usepackage{thmtools}
\usepackage{thm-restate}

\newtheorem{theorem}{Theorem} 
\newtheorem{informal_theorem}{Theorem} 
\newtheorem{corollary}{Corollary} 
\newtheorem{lemma}{Lemma} 
\newtheorem{assumption}{Assumption} 
\newtheorem{definition}{Definition} 
\newtheorem{remark}{Remark}

\def\eg{\textit{e.g.,}}
\def\ie{\textit{i.e.,}}

\title{An Exponential Improvement on the Memorization Capacity of Deep Threshold Networks}

\newcommand*{\email}[1]{%
    \normalsize\href{mailto:#1}{\texttt{#1}}
    }

\author{\normalsize \ \ \quad \textbf{Shashank Rajput}\\ \normalsize \ \ \ \quad University of Wisconsin-Madison\\ \normalsize \ \ \quad \email{rajput3@wisc.edu}\and \normalsize\textbf{Kartik Sreenivasan} \\\normalsize University of Wisconsin-Madison\\ \normalsize\email{ksreenivasa2@wisc.edu} \and \normalsize\textbf{Dimitris Papailiopoulos}\\
\normalsize University of Wisconsin-Madison\\
\normalsize \email{dimitris@papail.io} \and \normalsize \quad \ \ \ \textbf{Amin Karbasi}\\\normalsize \quad \ \ \ Yale University\\ \normalsize \quad \ \ \ \email{amin.karbasi@yale.edu}}

\date{}
\begin{document}
\maketitle

\begin{abstract}
    It is well known that modern deep neural networks are powerful enough to memorize datasets even when the labels have been randomized. Recently, \cite{vershynin2020memory} settled a long standing question by \cite{baum1988capabilities}, proving that \emph{deep threshold} networks can   memorize $n$ points in $d$ dimensions using $\widetilde{\cO}(e^{1/\delta^2}+\sqrt{n})$ neurons and $\widetilde{\cO}(e^{1/\delta^2}(d+\sqrt{n})+n)$ weights, where $\delta$ is the minimum distance between the points. In this work, we improve the dependence on $\delta$ from exponential to almost linear, proving that $\widetilde{\cO}(\frac{1}{\delta}+\sqrt{n})$ neurons and $\widetilde{\cO}(\frac{d}{\delta}+n)$ weights are sufficient. Our construction uses Gaussian random weights only in the first layer, while all the subsequent layers use binary or integer weights. We also prove new lower bounds by connecting memorization in neural networks to the purely geometric problem of separating $n$ points on a sphere using hyperplanes.
\end{abstract}

\section{Introduction}
The current paradigm of training neural networks is to train the networks until they fit the training dataset perfectly \citep{Zhang17understanding, arpit2017closer}. This means that the network is able to output the exact label for every sample in the training set, a phenomenon known as \emph{interpolation}. Quite interestingly, modern deep networks have been known to be powerful enough to interpolate even randomized labels \citep{Zhang17understanding, liu2020bad}, a phenomenon that is usually referred to as \emph{memorization} \citep{yun2018small,vershynin2020memory,bubeck2020network}, where the networks can interpolate any arbitrary labeling of the dataset.

Given that memorization is a common phenomenon in modern deep learning, a reasonable question to ask is that of how big a neural network needs to be so that it can memorize a dataset of $n$ points in $d$ dimensions. This question has been of interest since the 80s \citep{baum1988capabilities,mitchison1989bounds,sontag1990remarks,huang1991bounds,sartori1991simple}. In particular, \cite{baum1988capabilities} proved that a single hidden layer threshold network with $\cO(\max(n,d))$ weights and $\cO(\lceil n/d\rceil)$ neurons can memorize any set of $n$ points in $d$ dimensions, as long as they are in general position.\footnote{The ``general position'' assumption of $n$ vectors in $\bR^d$ means that any collection of $d$ of these $n$ vectors are linearly independent.} 
Baum asked if going deeper could increase the memorization power of threshold networks, and in particular if one could reduce the number of neurons needed for memorization to $\cO(\sqrt{n})$. 
While for ReLU activated networks, \cite{yun2018small} were able to prove that indeed only $\cO(\sqrt{n})$ neurons were sufficient for memorization using deeper networks, the same question remained open for threshold networks.

Recently, \cite{vershynin2020memory} was able to answer Baum's question in the positive by proving that indeed $\widetilde{\cO}(e^{1/\delta^2}+\sqrt{n})$ neurons were sufficient to memorize a set of $n$ points on a unit sphere separated by a distance at least $\delta$. 
Many recent works study the memorization power of neural networks under separation assumptions, \eg~\cite{bubeck2020network,vershynin2020memory,park2020provable}. One reason why before Vershynin's work, it was unclear whether going deeper was helpful for threshold networks was that unlike ReLU or sigmoid functions, the threshold activation function, \ie{} $\sigma(z)={\bf 1}_{z\geq 0}$, prohibits neurons from passing `amplitude information' to the next layer.

Although the dependence of $\cO(\sqrt{n})$ on the number of neurons was achieved by the work of Vershynin, it is unclear that the exponential dependence on distance and the requirement that the points lie on a sphere is fundamental.
In this work, we lift the spherical requirement and offer an exponential improvement on the dependence on $\delta$.

\begin{informal_theorem}(Informal)
 There exists a threshold neural network with $\widetilde{\cO}\left( \frac{1}{\delta} + \sqrt{n}\right)$ neurons and $\widetilde{\cO}\left(\frac{d}{\delta} + n \right)$ weights that can memorize any $\delta$-separated dataset of size $n$ in $d$ dimensions.
\end{informal_theorem}
Please see Definition~\ref{def:delta_separation} for the formal definition of $\delta$-separation; informally, it means that all the points have bounded norm and a distance of at least $\delta$ between them. Comparing the theorem above with Vershynin's result, we see that firstly we have reduced the dependence on $\delta$ from exponential to nearly linear; and secondly, in our upper bound, the $\delta$ and $n$ terms appear in summation rather than product (ignoring the logarithmic factors). Further, note that \cite{vershynin2020memory} needs the points to lie on a sphere, whereas we only need them to have a bounded $\ell^2$ norm. We have compared our results with the existing work in Table~\ref{tab:comparison}.

\begingroup
\renewcommand{\arraystretch}{1.5}
\begin{table}[t]
{
\centering
\resizebox{\textwidth}{!}{%
\begin{tabular}{|l|l|l|l|}
\hline
\textbf{Reference} & $\#$ \textbf{Neurons} & $\#$ \textbf{Weights} & \textbf{Assumptions} \\ \hline
         \cite{baum1988capabilities} &     $\cO(\lceil {n}/{d}\rceil)$    &   $\cO(\max(n,d))$     &       General Position    \\ \hline
         \cite{huang1991bounds} &   $\cO(n)$      &   $\cO(nd)$      &        None     \\ \hline
                  \cite{vershynin2020memory} &     $\cO(e^{1/\delta^2}\log^p (n) + \sqrt{n}\log^{2.5} (n))$    &   $\cO(e^{1/\delta^2}(d +\sqrt{n})+n\log^5(n))$    &      $\delta$-separation, points lie\\& & &on the unit sphere       \\ \hline
         \textbf{Ours, Theorem~\ref{thm:upperBound}} &   $\cO\left(\dfrac{\log^2 n}{\delta}+\sqrt{n}\log^2 n\right)$      &  $\cO\left(\dfrac{(d+\log n)\log n}{\delta}+ n\log^2 n\right)$     &    $\delta$-separation    \\ \hline
\end{tabular}%
}
\caption{Table comparing the upper bounds for the parameters needed for a threshold network to memorize a dataset. General position assumption in $\bR^d$ means that no more than $d$ points lie on a $d-1$ dimensional hyperplane. Note: $\log^p (n)$ denotes a poly-log factor in $n$.}
\label{tab:comparison}
}
\end{table}
\endgroup

Our construction has $\cO(\log \frac{\log n}{\delta})$ layers, of which only the first layer in our construction has real weights, which are i.i.d Gaussian. This layer has $\cO(\frac{\log n}{\delta})$ neurons and $\cO(\frac{d\log n}{\delta})$ weights. All the other layers have integer weights which are bounded and are on the of order $\cO(\log n)$. 

\cite{baum1988capabilities} also proved that there exists a dataset of $n$ points such that any threshold neural network would need at least $\lceil n/d \rceil$ neurons in the first layer to memorize it. However, the distance for this dataset is $\cO(1/n)$.\footnote{Here, we have rescaled the dataset to have maximum norm of any sample to be 1. This is done to make the dataset consistent with our minimum distance definition (see Assumption~\ref{assumption:2}). The assumption in \cite{park2020provable} also uses this kind of normalization.}
Our upper bound (Theorem~\ref{thm:upperBound}) shows that if $\delta = \varOmega(1/n)$, then we can memorize with only $\cO(\frac{\log n}{\delta})$ neurons in the first layer. To complement our upper bound, we introduce a new, $\delta$-dependent lower bound below: 

\begin{informal_theorem}(Informal)
There exists a $\delta$-separated dataset of size $n\in \left[\frac{d^2}{\delta},\left(\frac{1}{\delta}\right)^{\frac{d}{2}}\right]$ such that any threshold network that can memorize it needs $\widetilde{\varOmega}\left(\frac{1}{\sqrt{\delta}}\right)$ neurons in the first layer.
\end{informal_theorem}

The rest of the paper is divided into 7 sections. In Section~\ref{sec:relWorks}, we provide the related works for memorization using neural networks. We provide definitions and notation in Section~\ref{sec:preliminaries} and our main results in Section~\ref{sec:main_results}. 
Then, we briefly explain the constructions of our upper bound in  Section~\ref{sec:upper_bound}.
Before exploring lower bounds for threshold networks, we provide sharp bounds on the minimum parameterization needed for any model (not necessarily a neural network) to memorize a $\delta$-separated dataset, in Section~\ref{sec:info_theory}. We discuss our lower bound for threshold networks in Section~\ref{sec:lower_bound}. Finally, we conclude our paper with a brief conclusion in Section~\ref{sec:conclusion}.

\section{Related works}\label{sec:relWorks}
 \paragraph*{Memorization with threshold activation.} There is a rich history of research into the memorization capacity of threshold networks. \cite{baum1988capabilities} showed that a single hidden layer network with $\cO(\max(n,d))$ weights and $\cO(\lceil n/d \rceil)$ neurons are sufficient to memorize a dataset, \emph{if} the samples are in general positions. Baum also provided an instance of a dataset, where any threshold based network aiming to memorize the dataset would need at least $\lceil n/d\rceil$ neurons in the first layer. However, the construction was such that $\delta$ scaled inversely with $n$. In this work, we show that for any minimum distance $0<\delta\leq \frac{1}{2}$, the number of neurons in the first layer can be much smaller. Baum also proved, using the Function Counting Theorem \citep{cover1965geometrical}, that $\cO(n/\log n)$ connections are needed in any neural network that aims to memorize points in general position, regardless of the minimum distance.

\cite{huang1991bounds} and \cite{sartori1991simple} proved that threshold networks could memorize any arbitrary set of points, that is, the points need not be in general position. However, these results needed $(n-1)d$ weights and $n-1$ neurons in the networks for memorization. \cite{mitchison1989bounds} and \cite{kowalczyk1997estimates} provided upper and lower bounds when the requirement of memorizing the dataset exactly was relaxed to allow a recall error of up $50\%$. \cite{sontag1990remarks} provided evidence that suggested that adding skip connections can double the memorization capacity of single hidden layer networks. \cite{kotsovsky2020bithreshold} extended Baum's result to multiclass classification setting, using bithresholding activation functions.

Recently, \cite{vershynin2020memory} was able to show that $\cO(e^{1/\delta^2}(d+\sqrt{n})+n)$ weights and $\cO(e^{1/\delta^2}+\sqrt{n})$ neurons are sufficient to memorize a set of $n$ points on the unit sphere, separated by a distance of at least $\delta$. This answered the question by \cite{baum1988capabilities}, which was whether the number of neurons could be reduced to $\cO(\sqrt{n})$ by considering deeper networks. Note that Veshynin's and Baum's assumptions are fundamentally different and neither is stronger than the other. For example, Baum's construction would still need only $\cO(\max(n,d))$ weights even if the points are infinitesimally close to each other, whereas Vershynin's construction would grow as $\delta$ decreases. On the other hand, Vershynin's construction works even if all the points lie on a low dimensional hyperplane, whereas Baum's general position assumption would get violated in this situation. Note that na\"{i}vely extending Baum's construction to the case where the points lie on a very low dimensional hyperplane would result in $\cO(nd)$ weights and $\cO(n)$ neurons, which is identical to the bounds by \cite{huang1991bounds}.

\paragraph*{Memorization with ReLU and sigmoid activation.} Due to the popularity of the ReLU activation function, there has also been an interest in their memorization capacity. \cite{Zhang17understanding} proved that a single hidden layer ReLU network with $2n + d$ weights and $n$ neurons can memorize $n$ arbitrary points. \cite{hardt2017identity} proved that ReLU networks with residual connections also need only $\cO(n)$ weights to memorize $n$ points. \cite{yun2018small} were able to prove that for ReLU, in fact, one needs only $\cO(\sqrt{n})$ neurons to memorize $n$ points. Quite interestingly, the construction by \cite{vershynin2020memory} works for both threshold activated networks (as described above) as well as ReLU activated networks; and gives the same bounds for both architectures. \cite{bubeck2020network} show that under some minimum distance assumptions, ReLU networks can memorize (even real labels) with $\cO(n/d)$ neurons such that the total weight \citep{bartlett1998sample} of the network is $\cO(\sqrt{n})$, which is proved to be optimal for single hidden layer networks. \cite{huang2003learning} proved that a two-hidden layer network with $\cO(\sqrt{n})$ sigmoid activated neurons suffices to memorize $n$ points. Recently, \cite{park2020provable} have shown that ReLU or sigmoid activated networks with only $\cO(n^{2/3}+\log(1/\delta))$ weights are sufficient to memorize $n$ points separated by a normalized distance of $\delta$.

\paragraph*{Over parameterization and generalization.} A large body of recent work attempts to explain why large neural networks, which are capable memorizers, can also generalize well, \eg{}  \cite{neyshabur2018role,neyshabur2019towards, belkin2020two}. In this work, we only study memorization, and omit any discussion about generalization. We do not believe our results  explain anything about generalization.

\section{Preliminaries}\label{sec:preliminaries}

In this work, we will consider feed forward neural networks $f:\bR^d\to \{0,1\}$ of the form
\begin{align*}
    f(\vx):=\sigma(b_L+\vw_L^\top\sigma(\vb_{L-1}+\mW_{L-1}\sigma(\vb_{L-2}+\mW_{L-2}\sigma(\dots \sigma(\vb_{1}+\mW_{1}\vx) \dots)))),
\end{align*}
where $\sigma(\cdot)$ is the threshold activation. The network has $L$ layers, with the last layer consisting of a single neuron with bias $b_L\in \bR$, and weight vector $\vw_L$. For any other layer $l$, its bias vector is $\vb_l$ and weight matrix is $\mW_l$. We use the notation $[n]$ to denote the set $\{1,\dots, n\}$.

We formally define memorization as follows:
\begin{definition}
A learning algorithm $\cA$ can memorize a dataset of feature vectors $\cD=\{\vx_i\}_{i=1}^n$ if for any arbitrary labeling $\{y_i\}_{i=1}^n$ of the dataset, the algorithm can output a model $f$ such that $\forall i\in [n]: f(\vx_i)=y_i$.
\end{definition}
For instance, $\cA$ could be the training procedure for a particular neural architecture and $f$ could be the neural network that $\cA$ outputs after the training process. For neural networks, we will abuse the notation and say that a neural network $f$ can memorize a dataset $\cD=\{\vx_i\}_{i=1}^n$ if for every arbitrary labeling $\{y_i\}_{i=1}^n$ of the dataset, we can find a set of weights and biases for the network so that $\forall i\in [n]: f(\vx_i)=y_i$.

We are mainly interested in the minimum size of a threshold network so that it can memorize a dataset of $n$ samples. Our only assumption on the dataset will be that of $\delta$-separation, which we define below.
\begin{definition}\label{def:delta_separation}
We say that a dataset is $\delta$-separated if it satisfies either Assumption~\ref{assumption:1} \textbf{or} Assumption~\ref{assumption:2}.
\end{definition}
\begin{assumption}(Angular separation)\label{assumption:1}
All the feature vectors in the dataset, $\{\vx_i\}_{i=1}^n$ satisfy 
\begin{align*}
\arccos\left(\frac{\langle \vx_i,\vx_j\rangle}{\|\vx_i\|\|\vx_j\|}\right)\geq \delta 
\end{align*}
\end{assumption}
\begin{assumption}(Normalized minimum distance)\label{assumption:2}
All the feature vectors in the dataset, $\{\vx_i\}_{i=1}^n$ satisfy $\|\vx_i\|\leq 1$ and 
\begin{align*}
\forall i\neq j&: \|\vx_i-\vx_j\|\geq \delta.
\end{align*}
\end{assumption}

\begin{remark}
Note that the `normalized' in the name of Assumption~\ref{assumption:2} refers to the fact that the maximum norm of any point in the dataset is bounded by 1. Our results can be easily extended to any other bounded dataset by rescaling the weights and biases of the first layer appropriately. \cite{park2020provable} also use a similar normalized distance assumption.
\end{remark}

Note that if the points lie on the unit sphere, like in \cite{vershynin2020memory} or Theorem~\ref{thm:lowerBound} in this paper, then both assumptions are roughly equivalent.

\textbf{Notation:} We use the lower case letters for scalars ($w$), lower case bold letters for vectors ($\vw$), and upper case bold letters for matrices ($\mW$). We denote the $i$-th element of vector $\vw$ as $w_i$, similarly for the case when the vectors have subscript, we denote the $i$-th element of $\vw_j$ by $w_{j,i}$. We use a tilde over $\cO(\cdot)$, $\varOmega(\cdot)$ and $\varTheta(\cdot)$, \ie{} $\widetilde{\cO}(\cdot)$, $\widetilde{\varOmega}(\cdot)$ and $\widetilde{\varTheta}(\cdot)$, to hide the logarithmic factors.

\section{Main results}\label{sec:main_results}
As discussed earlier, the existing result by \cite{vershynin2020memory} has an exponential dependence on $\delta$. In this section, we provide a new and tighter upper bound on the number of weights and neurons needed for a memorization, which brings the dependence down to almost linear. Our result is stated below:
\begin{theorem} \label{thm:upperBound}
There exists a threshold activated neural network with $\cO( \frac{\log^2 n}{\delta} + \sqrt{n}\log^2 n)$ neurons and\\ $\cO\left(\frac{(d+\log n)\log n}{\delta} + n\log^2 n\right)$ weights that can memorize any dataset of size $n$ in $d$ dimensions that is $\delta$-separated.
\end{theorem}

In this construction, only the first layer has real weights. The rest of the network has binary and integer weights. In particular, the $O(n\log^2 n)$ term in the number of weights comes from binary and integer weights (with the integers being bounded and on the order of $
\cO(\log n)$). 
Hence, the $\cO(n\log^2 n)$ weights can actually be represented by $\cO(n\log^2 n \log \log n)$ bits. 
We will see in Section~\ref{sec:info_theory} that $\Omega(n)$ bits are necessary for memorization by any model and hence this term is optimal, up to logarithmic factors. The $\frac{d\log n}{\delta}$ weights in the bound are the weights of the first layer which has width $\frac{\log n}{\delta}$, for which later, in Theorem~\ref{thm:lowerBound}, we will provide some indication that this factor might also be necessary.

\cite{baum1988capabilities} had constructed a dataset with normalized minimum distance $\cO(1/n)$ and proved that for this dataset, any threshold network aiming to memorize this must have $\lceil n/d\rceil$ neurons. We introduce a new, $\delta$-dependent lower bound below:
\begin{theorem}\label{thm:lowerBound}
For every $0<\delta\leq \frac{1}{2}$ and some universal constants $C_1$ and $C_2$, there exists a dataset of size $n\in \left[\frac{C_1d^2\log^2(d/\delta)}{\delta},\left(\frac{C_2}{\delta}\right)^{\frac{d}{2}}\right]$ satisfying the $\delta$-separation condition such that any threshold network that can memorize it needs $\varOmega\left(\frac{1}{\sqrt{\delta}\log (1/\delta)}\right)$ neurons in the first layer.
 \end{theorem}
Note that the dataset in this theorem actually satisfies \emph{both} Assumption~\ref{assumption:1} and Assumption~\ref{assumption:2}, up to a constant factor. To understand the implication of Theorem~\ref{thm:lowerBound} a bit better, consider two dataset $\cD_1$ and $\cD_2$, both containing $n$ points on the sphere. We assume that $\cD_1$ has $\delta_1$-separation and $\cD_2$ has $\delta_2$-separation, where $\delta_1=\varTheta(n^{-2/d})$ and $\delta_2=\widetilde{\varTheta}(d^2/n)$. Then, Theorem~\ref{thm:upperBound} says that there exists a network that can memorize $\cD_1$ with only $\cO({n^{2/d}}{\log n})$ neurons in the first layer, whereas Theorem~\ref{thm:lowerBound} says that any network aiming to memorize $\cD_2$ would need at least $\widetilde{\varOmega}(\sqrt{n}/d)$ neurons in the first layer. Hence, we see that the minimum distance can have a big impact on the network architecture.

\section{Memorization with threshold networks}\label{sec:upper_bound}

\begin{figure}[t]
    \centering
    \includegraphics[width=\textwidth]{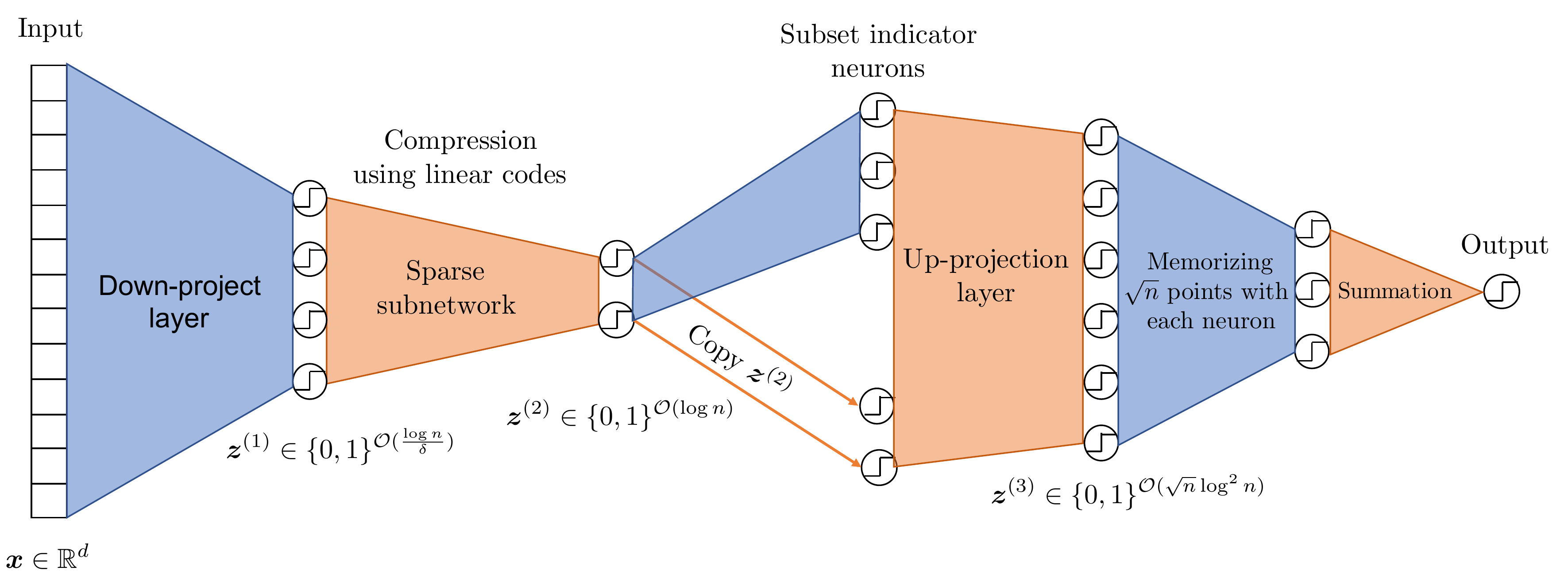}
    \caption{The schematic of the neural network that can memorize any $\delta$-separated dataset of size $n$ in $d$ dimensions (Theorem~\ref{thm:upperBound}). The first layer has random Gaussian weights and projects the input from $\bR^{d}$ down to $\{0,1\}^{\cO(\frac{\log n}{\delta})}$. The next component is a sparse subnetwork that further compresses the vectors from $\{0,1\}^{\cO(\frac{\log n}{\delta})}$ to $\{0,1\}^{\cO({\log n})}$ using linear codes. The next couple of layers lift the vectors from $\{0,1\}^{\cO({\log n})}$ to $\{0,1\}^{\cO(\sqrt{n}\log^2 n)}$ while ensuring that we can find sets of vectors of size $\sqrt{n}$ that are linearly independent. Finally, we use this linear independence to memorize up to $\sqrt{n}\log^2 n$ points with a single neuron. This gives us a layer of size $\sqrt{n}$ that can memorize the entire dataset. The final layer is just a single neuron that sums the outputs of these $\sqrt{n}$ neurons.}
    \label{fig:network_schematic}
\end{figure}

In this section, we give a proof sketch of Theorem~\ref{thm:upperBound}, the complete proof is provided in the appendix. We will construct a network that can memorize any $\delta$-separated dataset. As we discussed before, the threshold activation prohibits the passing of amplitude information to deeper layers. This is the biggest obstacle in leveraging the benefits of multiple layers of transformation that can be obtained by networks with activations such as ReLU. In the following, we will explain our construction layer-wise:

\paragraph*{Step 1: Generating unique binary representations.} Given that the amplitude information is lost after thresholding, we want to ensure that at the least, the first layer is able to transform the inputs into binary representations such that each sample in our dataset has a unique binary representation. Later, we will see that this is sufficient for memorization. 

Now we give an overview of how a layer with $\cO(\frac{\log n}{\delta})$ neurons with random i.i.d. weights can convert the input into unique binary vectors of length $\cO(\frac{\log n}{\delta})$. In this sketch, we will do this under Assumption~\ref{assumption:1}, but the same technique can be extended for Assumption~\ref{assumption:2}.

The key result we use is that any hyperplane passing through the origin, with random Gaussian coefficients has at least a $\delta/2\pi$ probability of separating two points which have an angle of at least $\delta$ between them. To see how this is true, consider two points $\vx_i$ and $\vx_j$ with an angle $\delta$ between them. Consider the 2-dimensional space spanned by these two vectors. Note that the intersection of the Gaussian hyperplane with the 2-dimensional space spanned by $\vx_i$ and $\vx_j$ is just a line passing through the origin. Further, because the Gaussian hyperplane has isometric coefficients, it can be shown that the line has angle uniformly distributed in $[0,2\pi)$. This implies that the probability that this line passes in between the two points is at least $\delta/2\pi$.
\begin{wrapfigure}[16]{r}{0.3\textwidth}
    \centering
    \centerline{\includegraphics[width=0.28\columnwidth]{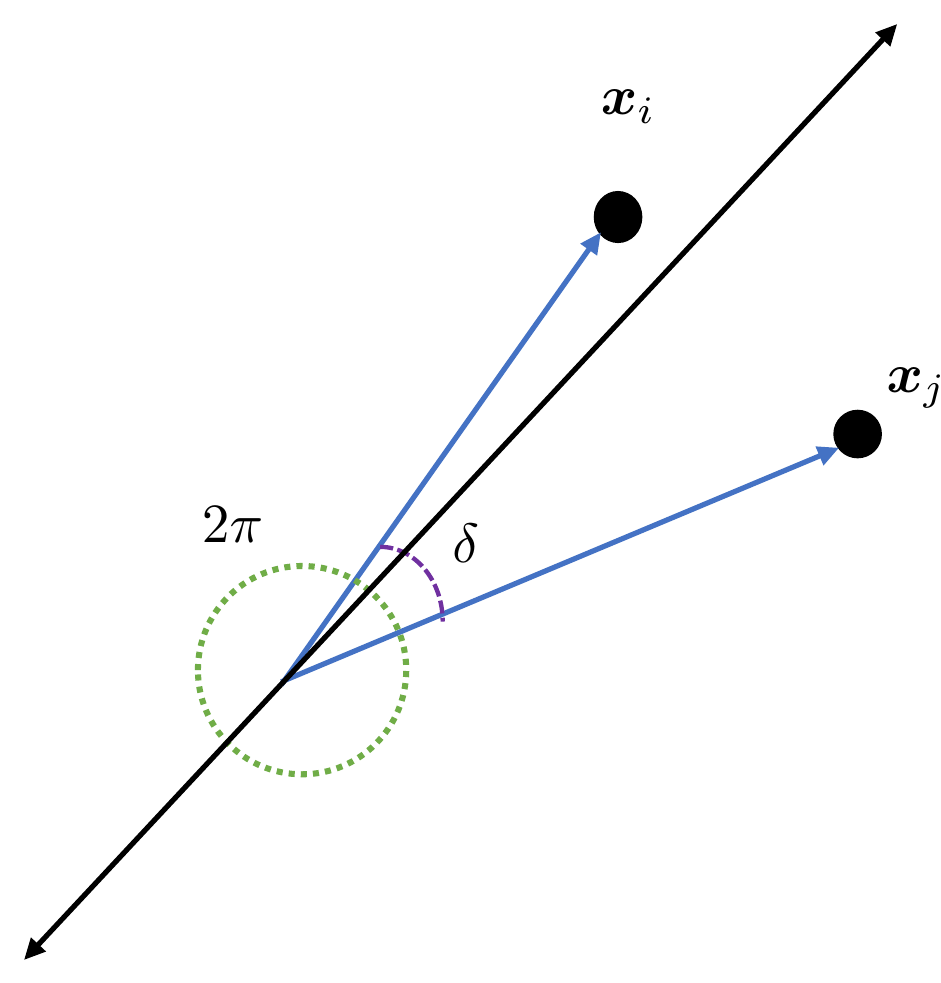}}
    \caption{The probability that a uniformly random line passing through the origin passes in between $\vx_i$ and $\vx_j$ is $\delta/2\pi$.}
    \label{fig:delta_cone}
\end{wrapfigure}
If we take $m$ such independent Gaussian hyperplanes, then the probability that none of them separate a pair $\vx_i,\vx_j$ is less than $(1-\frac{\delta}{2\pi})^m$. Finally, taking a simple union bound over the $\binom{n}{2}$ pair of points, shows that the probability that at least one pair has no hyperplane passing in between them is at most
\begin{align*}
    \binom{n}{2} \left(1-\frac{\delta}{2\pi}\right)^m.
\end{align*}
We want this to be less than $1$ to show that there exists one set of hyperplanes which can separate all the pairs. Doing that gives us $m = \cO(\frac{\log n}{\delta})$, which is our required bound.

For any input $\vx_i$, each threshold neuron in the first layer outputs either 0 or 1. We can concatenate the outputs of all the neurons in the first layer into a binary vector. Let us denote this binary representation of $\vx_i$ created by the first layer as $\vz_i^{(1)}$, for $i=1,\dots, n$. Further, denote the dimension of $\vz_i^{(1)}$ by $d^{(1)}$, which is equal to the number neurons in the first layer, that is,  $d^{(1)}=\cO(\frac{\log n}{\delta})$.

\paragraph*{Step 2: Compressing the binary representations to optimal length.} In order to avoid a factor $\delta$ in the later layers, our next task is to further compress the $\cO\left(\frac{\log n}{\delta}\right)$ dimensional vectors returned by the first layer into vectors of length $\cO(\log n)$. Note that $\cO(\log n)$ length is optimal (up to constant multiplicative factors) if we want to represent the $n$ samples uniquely using binary vectors. For this we will use linear codes where we simulate XOR operations by threshold activations.

Consider two $d^{(1)}$-dimensional binary vectors $\vz_i^{(1)}$ and $\vz_j^{(1)}$ that differ in at least one bit. Let $\oplus$ represent the XOR operation. Let $\vb$ be a random binary vector of length $d^{(1)}$ where each bit is i.i.d Bernoulli(0.5). Then, we claim that 
\begin{align*}
    \Pr(\langle \vz_i^{(1)},\vb\rangle_{\oplus}\neq \langle \vz_j^{(1)},\vb\rangle_{\oplus})&=0.5,\\
\text{where } \langle \vz ,\vb\rangle_{\oplus} &= (z_{1}\cdot b_{1})\oplus \dots \oplus (z_{d^{(1)}}\cdot b_{d^{(1)}}) 
\end{align*}

For the ease of exposition, assume that $\vz_i^{(1)}$ and $\vz_j^{(1)}$ differ in only the first bit (the proof of the general case is provided in the appendix). Let the first bit of $\vz_i^{(1)}$ be 0, and that of $\vz_j^{(1)}$ be 1, and we have assumed that the rest of their bits are the same. Let $c:=(z_{i,2}^{(1)}\cdot b_{2})\oplus \dots \oplus (z_{i,d^{(1)}}^{(1)}\cdot b_{d^{(1)}})=(z_{j,2}\cdot b_{2})\oplus \dots \oplus (z_{j,d^{(1)}}\cdot b_{d^{(1)}})$. Then,
\begin{align*}
    \Pr(\langle \vz_i^{(1)},\vb\rangle_{\oplus}\neq \langle \vz_j^{(1)},\vb\rangle_{\oplus})&=\Pr(((0\cdot b_{1})\oplus c)\neq ((1\cdot b_{1})\oplus c)).
\end{align*}
It can now be verified that if $b_1=1$, then $((0\cdot b_{1})\oplus c)\neq ((1\cdot b_{1})\oplus c)$. Since $\Pr(b_1=1)=0.5$, we get the desired result.

We have shown that the probability that one such random vector $\vb$ differentiates between a pair $\vz_i^{(1)}$ and $\vz_j^{(1)}$ with probability $0.5$. Doing an analysis similar to the one that we did for the first layer, we get that $m=\cO(\log n)$ vectors suffice to separate all pairs. The big task here, however, is to implement the XOR operation using threshold activations. In the appendix, we show that this can in fact be achieved using a couple of sparse layers.

Similar to the output of the first layer, let us denote the compressed binary representation of $\vz_i^{(1)}$ created by these sparse layers as $\vz_i^{(2)}$, for $i=1,\dots, n$. Further, let $\vz_i^{(2)}$ have dimension $d^{(2)}$, where $d^{(2)}=\cO({\log n})$, as we showed above.

\paragraph*{Step 3: Partitioning the samples into subsets of size $\cO(\sqrt{n}\log n)$.}
Once we have $d^{(2)}=\cO({\log n})$ length binary representations of the inputs, memorizing with a single layer consisting of $(d^{(2)}+1)n$ weights and $n$ neurons is not difficult. However, our aim is to reduce the number of neurons to $\cO(\sqrt{n}\log n)$. For that, we use a strategy similar to \cite{yun2018small}: we will memorize up to $K$ samples with 1 neuron, where $K=\varOmega(\sqrt{n}\log n)$. Thus, we will need only $\cO(\sqrt{n})$ neurons in total.

\begin{wrapfigure}[30]{r}{0.40\textwidth}
    \centering
    \centerline{\includegraphics[width=0.40\columnwidth]{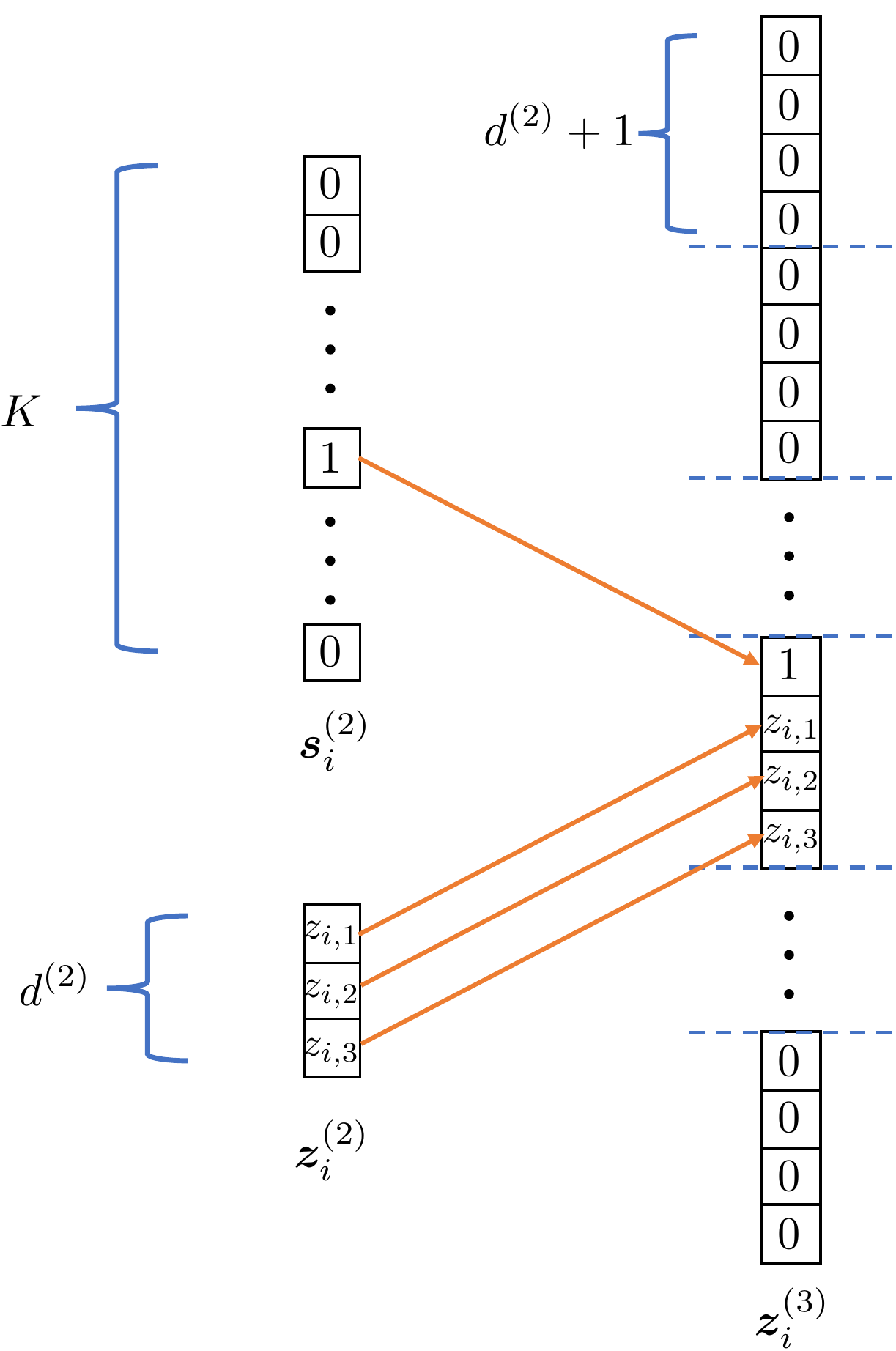}}
    \caption{In Step 3, $\vz_i^{(2)}$ is copied to a location in $\vz_i^{(3)}$ depending on the subset $S_j$ to which $\vz_i^{(2)}$ belongs.}
    \label{fig:up_project}
\end{wrapfigure}

Roughly speaking the strategy would be the following: Let's say we have a neuron $\sigma(\vw^\top \vz +b)$. Then, if for any $\{y_1, \dots,y_K\}\in \{0,1\}^K$, we can find a $\vw$ and $b$ such that all of the following equations hold at the same time, 
\begin{align}
    \vw^\top \vz_1 + b = y_1,\quad\dots\quad , \vw^\top \vz_K + b = y_K,\label{eq:lastLayerMemorize}
\end{align}
then this neuron can memorize the $K$ samples $\{\vz_1,\dots \vz_K\}$.

To do so, we need that the vectors $\{\vz_1,\dots,\vz_K\}$ be linearly independent. Since we have $K=\varOmega(\sqrt{n}\log n)$, we require that $\vz_i$'s have dimension at least $K$. However, the previous few layers have compressed the samples into $d^{(2)}=\cO(\log n)$ length binary vectors.

Thus, the next task is to project these $d^{(2)}$ length binary vectors to $\varOmega(K)$ length binary vectors. Note that first compressing down to $\cO(\log n)$ length vectors and then expanding these up to $\varOmega(\sqrt{n}\log n)$ length binary vectors seems counter intuitive. However, we were unable to project the vectors directly to $\varOmega(\sqrt{n}\log n)$ length binary vectors without incurring a large dependence on $\delta$ and $n$ in the number of weights and neurons.

One more objective that we accomplish while projecting up to $\varOmega(K)$ dimension is that both the projection and memorizing the resulting vectors can be easily done with bounded integer weights. The way we do this is by partitioning the $n$ vectors $\vz_i^{(2)}$ into $K$ subsets of size at most $\sqrt{n}$ each. We prove that we can find such $K$ subsets, such that each is characterized by a unique prefix, which all the elements of that subset share. These prefixes ensure that we can easily detect which subset a particular $\vz_i^{(2)}$ belongs to, using threshold neurons. For more details on this, please refer to the appendix.

Let $\{S_1,\dots,S_K\}$ denote the $K$ subsets, each with a unique prefix. Let $\vz_i^{(2)}\in\{0,1\}^{d^{(2)}}$ be the binary vector returned by \textbf{Step 2} above for input $\vx_i$. We mentioned above that by using threshold neurons, we can create a vector $\vs_i^{(2)}\in \{0,1\}^{d^{(2)}}$ for $\vz_i^{(2)}$ such that $s_{i,j}=1$ if and only if $\vz_i^{(2)}\in S_j$. Then, using these prefixes and threshold neurons, we can create the expanded binary vector $\vz_i^{(3)}\in\{0,1\}^{K(d^{(2)}+1)}$ corresponding to the vector $\vz_i^{(2)}$. The way we create $\vz_i^{(3)}$ from $\vz_i^{(2)}$ is as follows: $\vz_i^{(3)}$ is partitioned into $K$ chunks of length $d^{(2)}+1$. If $\vz_i^{(2)}\in S_j$, then the first bit of the $j$-th chunk is set to 1 and $\vz_i^{(2)}$ is copied to the rest of the bits in that chunk. All the other bits in $\vz_i^{(3)}$ are 0. This is shown in Figure~\ref{fig:up_project}. This operation can be done with threshold activated neuron, for details please refer to the appendix.

\paragraph*{Step 4: Memorizing the samples with $\sqrt{n}$ neurons}
A key property about the vectors $\vz_i^{(3)}$ is that if we take $K$ such vectors, each belonging to a different subset $S_j$, then they will be linearly independent as their coordinates do not overlap. 
Thus, we can find $\vw$ and $b$ such that \eqref{eq:lastLayerMemorize} is satisfied. 
Furthermore, due to the fact that these $K$ vectors have different supports, we can manage to satisfy \eqref{eq:lastLayerMemorize} with $\vw$ which have integer elements on the order of $\cO(\log n)$. 

Hence, we can memorize up to $K$ samples with one neuron. This gives that we need $\sqrt{n}$ in the second-to-last layer to memorize all the samples. The final layer is simply a single neuron that sums the outputs of the second-to-last layer. 

\section{Bit complexity of memorization}\label{sec:info_theory}
Before understanding the minimum size that threshold networks have to be in order to memorize a dataset, a more fundamental question is how many bits are needed to specify the models output by any learning algorithm that aims to memorize a dataset. In this section, we provide two theorems that sharply characterize the bit complexity needed for memorization. The results provide the bounds in the terms of two very related quantities, the \emph{packing number} and the \emph{covering number}, which we define below.

\begin{definition}
Given a real number $\delta >0$, the packing number $\cP_\delta$ of a set $\cS$ (equipped with norm $\|\cdot\|$) is defined as the cardinality of the largest set $\cX \subset \cS$ such that 
\begin{align*}
\forall x_1,x_2\in \cX: \|x_1-x_2\|\geq \delta.
\end{align*}
\end{definition}
Intuitively, the packing number is the maximum number of $\delta/2$ radius non-overlapping balls that one can fit inside a set.
\begin{definition}
Given a real number $\delta >0$, the covering number $\cC_\delta$ of a set $\cS$ (equipped with norm $\|\cdot\|$) is defined as the cardinality of the smallest set $\cX \subset \cS$ such that 
\begin{align*}
\forall x\in \cS, \exists \hat{x}\in \cX: \|x-\hat{x}\|\leq \delta.
\end{align*}
\end{definition}
Intuitively, the covering number is the minimum number of $\delta$ radius (possibly overlapping) balls that one needs to completely cover a set.

These quantities are very useful in quantifying the `size' of a set. These are related to each other through the following relation \citep[Lemma 4.2.8]{vershynin2018high}:
\begin{align*}
\cP_{2\delta} \leq \cC_\delta\leq\cP_{\delta}.
\end{align*}

Now, we are ready to state the upper and lower bounds on the bit complexity of memorization.
\begin{theorem}\label{thm:info_theoretic_lower_bound}
Let $\cS$ be a set from which we select the samples. Let $\cA$ be a learning algorithm that can memorize any dataset of feature vectors $\cD\in \cS^n$ as long as every pair of feature vectors $\vx_i,\vx_j\in \cD$ satisfies $\|\vx_i-\vx_j\|\geq \delta$. Then, we need at least $\max(n,\log_2 \log_2 \cP_\delta)$ bits to represent the models output by $\cA$.
\end{theorem}

\begin{theorem}\label{thm:info_theoretic_upper_bound}
Let $\cS$ be a set from which we select the samples. Then, there exists a learning algorithm $\cA$ that can memorize any dataset of feature vectors $\cD\in \cS^n$ as long as every pair of feature vectors $\vx_i,\vx_j\in \cD$ satisfies $\|\vx_i-\vx_j\|\geq \delta$, and the models returned by $\cA$ can be represented in $O(n+\log_2 \log_2 \cC_{\delta/2})$ bits.
\end{theorem}

Note that since $a+b=\cO(\max(a,b))$ for non-negative variables $a$ and $b$, the upper bound from Theorem~\ref{thm:info_theoretic_upper_bound} is of the same order as the lower bound from Theorem~\ref{thm:info_theoretic_lower_bound}.

It is known that both the packing and covering number of the unit sphere are of the order $\varTheta(1/\delta^d)$. Substituting this into Theorems~\ref{thm:info_theoretic_lower_bound} and \ref{thm:info_theoretic_upper_bound}, we get the following corollaries.

\begin{corollary}
Let $\cS^{d-1}$ be the unit sphere in $\bR^d$. Let $\cA$ be a learning algorithm that can memorize any set of $n$ points on $\cS^{d-1}$ as long as every pair of points $\vx_i,\vx_j\in \cD$ satisfies $\|\vx_i-\vx_j\|\geq \delta$. Then, we need $\varOmega(\max(n,\log d +\log  \log\frac{1}{\delta}))$ bits to represent the models output by $\cA$.
\end{corollary}

\begin{corollary}
Let $\cS^{d-1}$ be the unit sphere in $\bR^d$. Then, there exists a learning algorithm $\cA$ that can memorize any set of $n$ points on $\cS^{d-1}$ as long as every pair of feature vectors $\vx_i,\vx_j\in \cD$ satisfies $\|\vx_i-\vx_j\|\geq \delta$, and the models returned by $\cA$ can be represented in $O(n+\log d +\log  \log\frac{1}{\delta})$ bits.
\end{corollary}

Note that if we have a set of $n$ points on the unit sphere separated by at least distance $\delta$, then the set satisfies \emph{both} Assumption~\ref{assumption:1} and Assumption~\ref{assumption:2}. Hence, these two corollaries give bounds on the bit complexity of memorization under both the $\delta$-separation assumptions.

\section{Lower bound construction}\label{sec:lower_bound}
For our lower bound construction, we will have $n$ points on the unit sphere in $d$-dimensions. Each neuron in the first layer represents a hyperplane in $d$-dimensions. The main observation that we use is that for every pair of points $\vx_i$ and $\vx_j$ such that they have different labels, there should be a hyperplane passing in between them. Thus, we aim to lower bound the minimum number of hyperplanes needed for this to happen.

\begin{wrapfigure}[26]{r}{0.41\textwidth}
    \centering
    \centerline{\includegraphics[width=0.39\columnwidth]{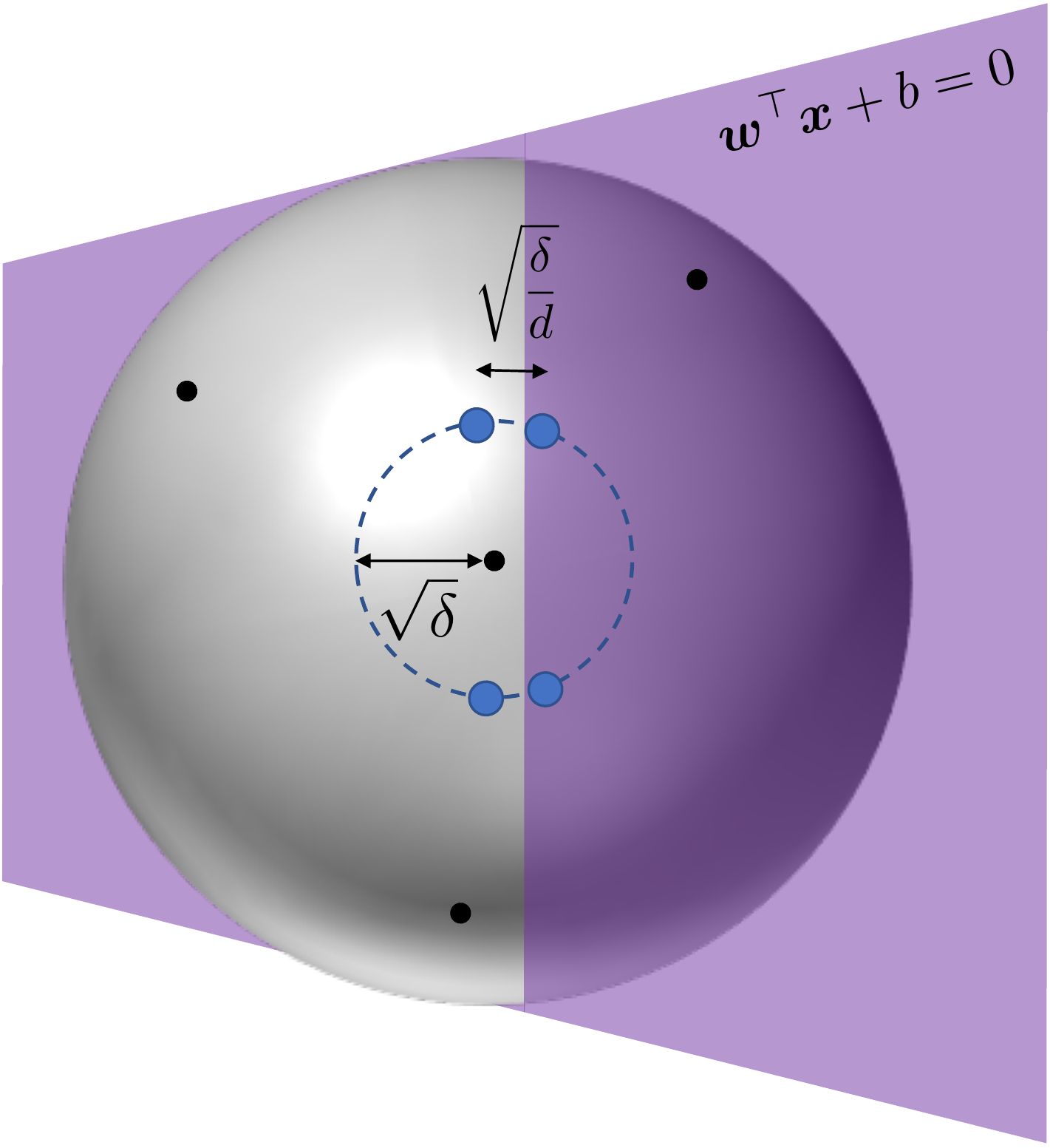}}
    \caption{In this figure, the black dots represent the cluster centers and the blue circles represent the samples in a cluster. In high dimensions, the points in each cluster get concentrated within distance $\cO(\sqrt{\delta/d})$ in any direction. Hence any hyperplane, that wants to separate a constant fraction of the points in a cluster, needs to pass within distance $\cO(\sqrt{\delta/d})$ of the cluster center, as shown in the figure.}
    \label{fig:lower_bound}
\end{wrapfigure}

This is related to the problem of having to separate every point using hyperplanes, that is, the problem where we need to ensure that there is a hyperplane passing in between \emph{every} pair of points $\vx_i$ and $\vx_j$. For this problem we have the following theorem.

\begin{theorem}\label{thm:hyperplaneSeparation}
For every $0<\delta\leq \frac{1}{2}$ and some universal constants $C_1$ and $C_2$, there exists a set of $n$ points on the $d$-dimensional sphere, with $n\in \left[\frac{C_1 d^2\log^5(d/\delta)}{\delta}, \left(\frac{C_2}{\delta}\right)^{\frac{d}{2}}\right]$, such that each pair of points is separated by a distance of at least $\delta$, and one needs $\varOmega\left(\frac{\log n}{\sqrt{\delta}\log \frac{\log n}{\delta}}\right)$ hyperplanes to separate them.
\end{theorem}

We use essentially the same construction for both the problems. On the sphere, first we sample $\sqrt{n}$ points uniformly at random, which we call \emph{cluster centers}.
Around each of these centers, we sample $\sqrt{n}$ points uniformly at random from the cap of radius $\sqrt{\delta}$, and call these sets of points \emph{clusters}. Hence, in total we sample $n$ points on the sphere and it can be shown that with high probability these will all be at least $\delta$ distance away from each other as long as $n\in \left[\frac{d^2}{\delta}, \frac{1}{\delta^{d/2}}\right]$. In high dimensions, it can be shown that for each cluster and in any direction, most points lie within a distance $\sqrt{\delta/d}$ of the cluster center. Hence, if a hyperplane aims to `effectively' separate pairs of points in a cluster, it needs to pass within distance $\sqrt{\delta/d}$ of the cluster center. It can also be shown that in high dimensions any hyperplane can only pass within distance $\sqrt{\delta/d}$ of the cluster centers of $\sqrt{\delta}$ fraction of the clusters. Finally, we use the following necessary inequality to get a bound on the required number of hyperplanes:
\begin{align*}
&(\#\text{ hyperplanes needed})  \\
&\ \geq \frac{(\#\text{ clusters})\times (\#\text{ hyperplanes needed per cluster})}{(\#\text{ clusters a hyperplane can `effectively' separate})}.
\end{align*}
We know that the number of clusters is $\sqrt{n}$, and we showed above that $(\#$ clusters a hyperplane can `effectively' separate$)$ is $\sqrt{\delta}$ fraction of the clusters, that is $\cO(\sqrt{n\delta})$. Finally, for the term $(\#\text{ hyperplanes needed per cluster})$, we use 1 for the memorization problem, where we need to separate points with opposite label. 
For the problem of separating every pair of points, we know that $(\#\text{ hyperplanes needed per cluster}) = \varOmega(\log_2 \sqrt{n})$ since there $\sqrt{n}$ points in each cluster and hence we need at least $\log_2 \sqrt{n}$ hyperplanes per cluster. 
Substituting these values gives us the bounds in Theorems \ref{thm:lowerBound} and \ref{thm:hyperplaneSeparation}.

Note that we skipped many details in the sketch above. The complete proof is provided in the appendix, and the construction there is slightly different.

\section{Conclusion}\label{sec:conclusion}
In this work, we study the memorization capacity of threshold networks under the $\delta$-separation assumption. We improve the existing bounds on the number of neurons and weights required, from exponential in $\delta$, to almost linear. We also prove new, $\delta$-dependent lower bounds for the memorization capacity of threshold networks, that together with our upper bound, shows that $\delta$ (the separation) indeed impacts the network architecture required for memorization.

\bibliography{references.bib}
\bibliographystyle{iclr2021_conference}

\newpage
\appendix

\input{appendix}

\end{document}

%% file: math_commands.tex
\usepackage{amsmath,amsfonts,bm}
\usepackage{bbm}
\usepackage[mathscr]{eucal}

\newcommand{\dPhi}{\textnormal{d}\phi}

\def\1{\bm{1}}

\def\rb{{\textnormal{b}}}

\def\rg{{\textnormal{g}}}

\def\rk{{\textnormal{k}}}

\def\rr{{\textnormal{r}}}

\def\rvb{{\mathbf{b}}}

\def\rvg{{\mathbf{g}}}

\def\rvw{{\mathbf{w}}}

\def\vZero{{\bm{0}}}
\def\vOne{{\bm{1}}}

\def\vb{{\bm{b}}}
\def\vc{{\bm{c}}}

\def\vs{{\bm{s}}}

\def\vu{{\bm{u}}}
\def\vv{{\bm{v}}}
\def\vw{{\bm{w}}}
\def\vx{{\bm{x}}}

\def\vz{{\bm{z}}}

\def\mI{{\bm{I}}}

\def\mW{{\bm{W}}}

\DeclareMathAlphabet{\mathsfit}{\encodingdefault}{\sfdefault}{m}{sl}
\SetMathAlphabet{\mathsfit}{bold}{\encodingdefault}{\sfdefault}{bx}{n}

\def\cA{{\mathcal{A}}}
\def\cB{{\mathcal{B}}}
\def\cC{{\mathcal{C}}}
\def\cD{{\mathcal{D}}}
\def\cE{{\mathcal{E}}}

\def\cH{{\mathcal{H}}}

\def\cN{{\mathcal{N}}}
\def\cO{{\mathcal{O}}}
\def\cP{{\mathcal{P}}}
\def\cQ{{\mathcal{Q}}}

\def\cS{{\mathcal{S}}}

\def\cX{{\mathcal{X}}}

\def\bF{{\mathbb{F}}}

\def\bR{{\mathbb{R}}}

%% file: appendix.tex
\section{Proof of Theorem~\ref{thm:upperBound}}
\begin{proof}
As explained in Section~\ref{sec:upper_bound}, the network memorizes in four steps. In this proof, we will go into the details of each step.

\paragraph*{Step 1: Generating unique binary representations.} The first layer's task is to generate unique binary representations for each sample in the dataset. 
What this means is that for every pair of samples $\vx_i$ and $\vx_j$, there should exist at least one neuron in the first layer, say $\sigma(\vw^\top \vx + b)$ such that $\sigma(\vw^\top \vx_i + b)\neq \sigma(\vw^\top \vx_j + b)$. 
Note that $\vw^\top \vx + b$ is just a hyperplane and $\sigma(\vw^\top \vx_i + b)\neq \sigma(\vw^\top \vx_j + b)$ is equivalent to saying that $\vx_i$ and $\vx_j$ lie on the opposite sides of the hyperplane. The next two lemmas says that we can easily find a set of $\cO(\log(n)/\delta)$ hyperplanes such that for every pair of points $\vx_i$, $\vx_j$, at least one hyperplanes passes in between them.
\begin{lemma}\label{lem:gaussian1}
Assume that our dataset satisfies Assumption~\ref{assumption:1}. Define $m:=\left\lceil\frac{4\pi}{\delta}\log \left(\frac{n}{\epsilon}\right)\right\rceil$ for any $\epsilon\in (0,1)$. If we sample $m$ hyperplanes of the form $\rvw_i^\top \vx =0$, for $i=1,\dots, m$, where $\rvw_i\stackrel{\text{i.i.d.}}{\sim}\cN(\vZero, \mI_d)$, then with probability $1-\epsilon$,
\begin{align*}
\forall i\neq j, \exists k: (\rvw_k^\top \vx_i)(\rvw_k^\top \vx_j)<0.
\end{align*}
\end{lemma}
\begin{lemma}\label{lem:gaussian2}
Assume that our dataset satisfies Assumption~\ref{assumption:2}. Define $m:=\frac{C}{\delta}\log \left(\frac{n}{\epsilon}\right)$ for some universal constant $C$ and any $\epsilon\in (0,1)$. If we sample $m$ hyperplanes of the form $\rvw_i^\top \vx +\rb_i=0$, for $i=1,\dots, m$, where $\rvw_i\stackrel{\text{i.i.d.}}{\sim}\cN(\vZero, \mI_d)$ and $\rb_i\stackrel{\text{i.i.d.}}{\sim}\cN(0, 1)$, then with probability $1-\epsilon$,
\begin{align*}
\forall i\neq j, \exists k: (\rvw_k^\top \vx_i+\rb_k)(\rvw_k^\top \vx_j+\rb_k)<0.
\end{align*}
\end{lemma}

Note that each neuron in the first layer outputs either 0 or 1. 
If we concatenate all the outputs of these $\cO(\log(n)/\delta)$ neurons, we will get a binary vector of length $\cO(\log(n)/\delta)$.
The lemmas above say that under Assumption~\ref{assumption:1} \emph{or} Assumption~\ref{assumption:2}, these binary vectors will be unique.

Let $\vz_i^{(1)}$ be the binary vector created by the first layer when the input to the network is $\vx_i$. Further, let $d^{(1)}$ denote the dimension (length) of this binary vector.
Note that the first layer has $\cO(\log(n)/\delta)$ neurons and $\cO(d\log(n)/\delta)$ weights and biases.

\paragraph*{Step 2: Compressing the binary representations.} The next few layers compress the $\cO(\log(n)/\delta)$ length binary vectors down to $\cO(\log(n))$ length unique binary representations for each sample. We will use linear codes in $\bF_2$ (the finite field of two elements: $0$ and $1$) for this.

For two binary vectors $\vb_1,\vb_2\in\{0,1\}^{d^{(1)}}$, define the inner product in the field as $$\langle \vb_1,\vb_2\rangle_{\oplus}:=(b_{1,1}\cdot b_{2,1})\oplus(b_{1,2}\cdot b_{2,2})\oplus\dots\oplus(b_{1,{d^{(1)}}}\cdot b_{2,{d^{(1)}}}),$$ where $\oplus$ is the XOR operator, and $b_{1,i}$ is the $i$-th element of $\vb_1$, and similarly $b_{2,i}$ is the $i$-th element of $\vb_2$.

Let $\rvb_1,\dots,\rvb_m$ be $m$ i.i.d. Bernoulli(0.5) vectors of length ${d^{(1)}}$ each. Define vectors $\vz_i^{(2)}$ as
\begin{align}
{\vz}_i^{(2)}=\begin{bmatrix}
\langle \rvb_1,\vz_i^{(1)}\rangle_{\oplus}\\
\vdots\\
\langle \rvb_m,\vz_i^{(1)}\rangle_{\oplus}
\end{bmatrix}, \forall i \in [n].\label{eq:gf2_inner_prod}
\end{align}

The next lemma says that we only need $m=\cO(\log n)$ to ensure that $\forall i\neq j$, $\vz_i^{(2)}\neq \vz_j^{(2)}$. 
\begin{lemma}\label{lem:GVLemma}
Let $\vz_1^{(1)},\dots,\vz_n^{(1)}$ be $n$ distinct binary vectors of dimension ${d^{(1)}}$, and let $\rvb_1,\dots,\rvb_m$ be $m$ vectors in  ${d^{(1)}}$ dimensions, with i.i.d. Bernoulli(0.5) entries. Let $\vz_1^{(2)},\dots,\vz_n^{(2)}$ be as defined in \eqref{eq:gf2_inner_prod}. If $m=3\log\frac{n}{\epsilon'}$, then with probability $1-\epsilon'$,
\begin{align*}
\forall i\neq j: \vz_i^{(2)}\neq \vz_j^{(2)}.
\end{align*}
\end{lemma}
The lemma above is a special case of the Gilbert-Varshamov bound (for example, see \citep{guruswami2012essential}). A simple proof of the lemma is provided in subsection~\ref{subsec:GVLemma} for completeness.

Hence, we will be able to compress the ${d^{(1)}}$-dimensional vectors down to $m$-dimensions. In particular this means we will be able to compress down from $\cO(\frac{\log n}{\delta})$ to $\cO(\log n)$ dimensions. \emph{However}, note that the key in the operation above is the XOR operator, so we will need to use threshold activation to simulate the XOR operation. The next lemma says that we can transform $\vz_i^{(1)}$ to $\vz_i^{(2)}$ using a threshold network containing $\cO(\frac{\log^2 n}{\delta})$ neurons and $\cO(\frac{\log^2 n}{\delta})$ weights.
\begin{lemma}\label{lem:thresh_to_xor}
Let $\vz_i^{(2)}$ be as defined in \eqref{eq:gf2_inner_prod}, for all $i\in[n]$. Then, we can construct a threshold network with $3d^{(1)}m$ neurons and $9d^{(1)}m$ weights and biases such that it outputs $\vz_i^{(1)}$ when the input to it is $\vz_i^{(2)}$, for all $i\in[n]$.
\end{lemma}

Hence, when the input to the overall network is $\vx_i$, the network constructed up till now will output $\vz_i^{(2)}$. Similar to the first layer, let $d^{(2)}=m$ denote the dimension of $\vz_i^{(2)}$. Note that this step uses $\cO(\frac{\log^2 n}{\delta})$ neurons and $\cO(\frac{\log^2 n}{\delta})$ weights.

\paragraph*{Step 3: Partitioning the dataset into $\widetilde{\cO}(\sqrt{n})$ subsets of size at most $\sqrt{n}$ each.} In this step, we partition $\{{\vz}_1^{(2)},\dots,{\vz}_n^{(2)}\}$ into $\cO(\sqrt{n}\log n)$ subsets of size at most $\sqrt{n}$ each. 
The reason why we do this will be apparent in \textbf{Step 4}.
The key property that we want these subsets to have is that each subset should have a unique prefix such that all the $\vz_i^{(2)}$ that belong to one subset share that prefix. 
This will help in detecting which subset a particular $\vz_i^{(2)}$ belongs to, using threshold networks.
Hence, first we will show that such prefixes exist.
\begin{lemma}\label{lem:prefix}
Given $n$ distinct binary vectors ${\vz}_1^{(2)},\dots,{\vz}_n^{(2)}$ of length $d^{(2)} = c \log_2 n$ each (for some constant $c\geq 1$), we can partition them into $\cO(\sqrt{n}\log n)$ subsets of size at most $\sqrt{n}$ each, with the important property that each of these subsets will be characterized by unique prefixes, such that all of the elements of one subset will have the same prefix.
\end{lemma}

Lemma~\ref{lem:prefix} gives us $K=\cO(\sqrt{n}\log n)$ subsets which partition $\{\vz_1^{(2)},\dots, \vz_n^{(2)}\}$, each with a unique prefix. Let the subsets be $S_1,\dots,S_K$. Let the binary prefix for $S_i$ be $\vb_i$, which a length $1\leq l \leq d^{(2)}$. Then the membership of a vector $\vz$ can easily be detected by the neuron $s_i:\{0,1\}^{d^{(2)}}\to \{0,1\}$ given by $s_i(\vz)=\sigma(\sum_{j=1}^l (2b_{i,j}-1)z_j - t)$, where $t$ is the number of 1's in $\vb_i$. 
Hence, we can have $K$ neurons, one for each subset, such that $s_{i}(\vz)=1$ if and only if $\vz\in S_i$. Let $\vs(\vz)$ represent the vector formed after concatenating the outputs of all these neurons. 
Along with these $K$ neurons, we will also have $d^{(2)}$ other neurons that simply copy $\vz^{(2)}$ over to the next layer. These $d^{(2)}$ neurons can simply have the form $f_i(\vz)=\sigma(2z_i - 1)$, for $i=1,\dots,d^{(2)}$. Overall, if $\vz_i^{(2)}\in S_j$, then this layer outputs
\begin{align}
\begin{bmatrix}
\vs(\vz_i^{(2)})\\
\vz_i^{(2)}
\end{bmatrix},\label{eq:indicator_layer}
\end{align}
as shown in the middle section of  Figure~\ref{fig:network_schematic}.
Thus, for \textbf{Step 3}, we have one layer which consists of $K$ subset indicator neurons, along with $d^{(2)}$ neurons that simply copy $\vz$ over to the output.  Overall, in this step we used one layer with $K+d^{(2)}$ neuron and $(K+1)d^{(2)}$ weights. 

\paragraph{Step 4:} The reason we divided the vectors into subsets and created the indicator neurons is so that the next layer can project the $d^{(2)}=\cO(\log n)$ dimensional vectors $\vz_i^{(2)}$ to $d^{(3)}=\cO(\sqrt{n}\log^2 n)$ dimension vectors using the subset indicators. Let $\vz_i^{(3)}$ be the projection of $\vz_i^{(2)}$. 
The way we project will ensure that if we take one vector from each of the $K$ subsets, then these vectors would be linearly independent. 
This linear independence of $K$ vectors would help in memorizing $K$ samples with just one neuron in the second-to-last layer, which means that we can memorize all the $n$ samples with $\sqrt{n}$ neurons in the second-to-last layer.
We explain all of this in detail now.

\paragraph*{Step 4(a): Projecting the vectors up to dimension $\cO(\sqrt{n}\log^2 n)$.}

In this layer, we will have $K(d^{(2)}+1)$ neurons partitioned into groups of size $d^{(2)}+1$ each. 
As explained in the main paper, the goal would be carry out the transformation shown in Figure~\ref{fig:up_project}.

To see how this happens, recall that the previous layer outputs \eqref{eq:indicator_layer}. 
Then, the $j$-th neuron of the $i$-th group in this layer would have the form  $g_{i,j}=\sigma(s_i - 1)$ for $j=1$ and $g_{i,j}=\sigma_{i,j}(s_i + z_j^{(2)} - 2)$, for $j=2,\dots, d^{(2)}+1$. It can be verified that this implements the transformation shown in Figure~\ref{fig:up_project}.

This layer will have $K(d^{(2)}+1)$ neurons and $\leq 3K(d^{(2)}+1)$ weights and biases. Further, looking at neurons, note that the weights and bias are integers and less than $2$.

\paragraph*{Step 4(b): Memorizing using $\sqrt{n}$ neurons.}
In this layer, we have $\sqrt{n}$ neurons. 
The $i$-th neuron memorizes the $i$-th samples from each of the $K$ subsets.
Let $\vz_{i,j}^{(2)}$ denote the $i$-th vector in $S_j$. 
The $i$-th neuron will have the weight vector given by
\begin{align*}
\vw^{(3)}_i=\begin{bmatrix}
-t_{i,1}+y_{i,1}-1\\
2\vz_{i,1}^{(2)} - \vOne\\
-t_{i,2}+y_{i,2}-1\\
2\vz_{i,2}^{(2)} - \vOne\\
\vdots\\
-t_{i,d^{(2)}}+y_{i,d^{(2)}}-1\\
2\vz_{i,d^{(2)}}^{(2)} - \vOne\\
\end{bmatrix},
\end{align*}
where $t_{i,j}$ is the number of $1$'s in $\vz_{i,j}^{(2)}$, and $y_{i,j}$ is the label of sample $\vx_{i,j}$ from the dataset that corresponds to $\vz_{i,j}^{(2)}$. The biases of these neurons are 0.

To see how this works, consider the vector $\vz_{i,j}^{(2)}$. This is the $i$-th vector in $S_j$. Then, the corresponding $\vz^{(3)}_{i,j}$ would have support only in the $j$-th chunk of length $d^{(2)}+1$. Then, 
\begin{align*}
   (\vw^{(3)}_i)^{\top} \vz_{i,j}^{(3)} &= -t_{i,j}+y_{i,j} - 1 + (2\vz_{i,j}^{(2)}-\vOne)^{\top} \vz_{i,j}^{(2)}\\
   &= -t_{i,j}+y_{i,j} - 1 + 2t_{i,j}-t_{i,j}\\
   &=y_{i,j} - 1.
\end{align*}
Hence, $\sigma ((\vw^{(3)}_i)^{\top} \vz_{i,j}^{(3)} )=y_{i,j}$. Further, for any $k\neq i$, 
\begin{align*}
   (\vw^{(3)}_i)^{\top} \vz_{k,j}^{(3)} &= -t_{i,j}+y_{i,j} - 1 + (2\vz_{i,j}^{(2)}-\vOne)^{\top} \vz_{k,j}^{(2)}.
\end{align*}
Note that $(2\vz_{i,j}^{(2)}-\vOne)^{\top} \vz_{j,j}^{(2)}\leq t_{i,j}-1$ if $\vz_{i,j}^{(2)}\neq \vz_{k,j}^{(2)}$. Hence, 
\begin{align*}
   (\vw^{(3)}_i)^{\top} \vz_{k,j}^{(3)} &\leq y_{i,j} - 2.
\end{align*}
Hence, $\sigma ((\vw^{(3)}_i)^{\top} \vz_{k,j}^{(3)} )=0$.

Thus we see that in this layer, the $i$-th neuron memorizes the $i$-th samples from each subset $\{S_1,\dots,S_K\}$, and for any other sample, it outputs $0$. Since there are at most $\sqrt{n}$ samples in each subset, we need $\sqrt{n}$ neurons in this layer.

Let $[q_1,\dots, q_{\sqrt{n}}]$ be the concatenated output of this layer. Then, the last layer is a single neuron, $\sigma(2q_1+\dots+2 q_{\sqrt{n}}-1)$, which essentially just sums up the outputs of the previous layer. Since each neuron of the previous layer memorized a subset of the samples, this neuron is able to output the correct label for the entire dataset.

\end{proof}
\subsection{Proof of helper lemmas for Theorem~\ref{thm:upperBound}}
\subsubsection{Proof of Lemma~\ref{lem:gaussian1}}
\begin{proof}
Consider any two samples from the dataset, $\vx_i$ and $\vx_j$. Let $\vu$ and $\vv$ be two orthonormal vectors that form a basis of the 2-dimensional space spanned by $\vx_i$ and $\vx_j$. Then, any point in the subspace can be written as $z_1\vu+z_2\vv$, for some scalars $z_1$ and $z_2$. Let $\vx_i = z_{i,1}\vu+z_{i,2}\vv$ and $\vx_j = z_{j,1}\vu+z_{j,2}\vv$.

As per our construction of the hyperplanes, let $\rvw^\top \vx =0$ be any one of the $m$ hyperplanes. Then, we want to find the probability that $(\rvw^\top \vx_i)(\rvw^\top \vx_j)<0$. For this, it would be sufficient that $\rvw^\top \vx_i>0$ and $\rvw^\top \vx_j<0$. These are equivalent to $z_{i,1}\rvw^\top \vu+z_{i,2}\rvw^\top\vv>0$ and  $z_{j,1}\rvw^\top \vu+z_{j,2}\rvw^\top\vv<0$. Note that since $\rvw\sim \cN(\vZero,\mI_d)$, we have that $\rvw^\top \vu\sim \cN(0,1)$ and $\rvw^\top \vv\sim \cN(0,1)$, and are independent. For convenience, define the random variables $\rg_1:=\rvw^\top \vu$, $\rg_2:=\rvw^\top \vv$, and $\rvg:= [\rg_1\quad \rg_2]^\top$. Also, define $\vz_i=[z_{i,1}\quad z_{i,2}]^\top $ and $\vz_j=[z_{j,1}\quad z_{j,2}]^\top $. Then, we want to bound the probability of event
\begin{align}
\{\rvg^\top \vz_i>0\ \text{and}
\ \rvg^\top \vz_j<0\}.\label{eq:lem1ineq}
\end{align}
Note that $\rvg,\vz_i$, and $\vz_j$ are all in 2 dimensions, the angle between $\vz_i$ and $\vz_j$ is at least $\delta$. Thus, for \eqref{eq:lem1ineq} to be true, $\rvg$ should lie in a cone of angle $\delta$ (see Figure~\ref{fig:delta_cone}). 
Noting that $\rvg$ is rotationally symmetric, we get that the probability of \eqref{eq:lem1ineq} is exactly $\frac{\delta}{2\pi}$. 
Hence, the probability that the hyperplane does not separate $\vz_i$ and $\vz_j$, is less than $1-\frac{\delta}{2\pi}$. Since we have $m$ independently sampled hyperplanes, the probability that none of them separates $\vz_i$ and $\vz_j$ is less than $(1-\frac{\delta}{2\pi})^m$. This is the probability that one of the pairs $\vx_i$ and $\vx_j$ does not get separated. Since there are a total of $\binom{n}{2}$ pairs of points $\vx_i$ and $\vx_j$, then using the union bound, the probability that there exists one pair of points such that the $m$ hyperplanes do not separate them, is less than $\binom{n}{2}(1-\frac{\delta}{2\pi})^m$. We want this probability to be less than $\epsilon$, that is,
\begin{align*}
&\binom{n}{2}\left(1-\frac{\delta}{2\pi}\right)^m \leq \epsilon.
\end{align*}
Note that $\binom{n}{2}\leq n^2$ and $\left(1-\frac{\delta}{2\pi}\right)^m\leq e^{-\frac{\delta}{2\pi}m}$. Hence, it is sufficient for the following to hold
\begin{align*}
n^2e^{-\frac{\delta}{2\pi}m} \leq \epsilon.
\end{align*}
This gives that $m\geq \frac{2\pi}{\delta}\log \left(\frac{n^2}{\epsilon}\right)$ is sufficient for the lemma to be true.
\end{proof}

\subsubsection{Proof of Lemma~\ref{lem:gaussian2}}
This proof is similar to the proof of Lemma~\ref{lem:gaussian1}.
\begin{proof}

Consider any two samples from the dataset, $\vx_i$ and $\vx_j$. Let $\vu$ and $\vv$ be two orthonormal vectors that form a basis of the 2-dimensional space spanned by $\vx_i$ and $\vx_j$, such that $\vv\perp (\vx_i-\vx_j)$.
If $\vx_i$ and $\vx_j$ do not span a 2-dimensional space, then take $\vu$ to be in the direction of $\vx_i$, and $\vv$ can be in any other orthogonal direction. 
Then, any point in the subspace can be written as $z_1\vu+z_2\vv$, for some scalars $z_1$ and $z_2$. Let $\vx_i = z_{i,1}\vu+z_{i,2}\vv$ and $\vx_j = z_{j,1}\vu+z_{j,2}\vv$. Note that because $\vv\perp (\vx_i-\vx_j)$, we have that $z_{i,2}=z_{j,2}$, and also $|z_{i,1}-z_{j,1}|=\|\vx_i-\vx_j\|\geq \delta$.

As per our construction of the hyperplanes, let $\rvw^\top \vx +\rb=0$ be any one of the $m$ hyperplanes. We are interested in the probability that $(\rvw^\top \vx_i+\rb_i)(\rvw^\top \vx_j+\rb_j)<0$, or equivalently 
\begin{align}
( z_{i,1}\rvw^\top\vu+ z_{i,2}\rvw^\top\vv+\rb)( z_{j,1}\rvw^\top\vu+z_{j,2}\rvw^\top\vv+\rb)<0.\label{eq:lem2ineq}
\end{align}

Note that since $\rvw\sim \cN(\vZero,\mI_d)$, we have that $\rvw^\top \vu\sim \cN(0,1)$ and $\rvw^\top \vv\sim \cN(0,1)$, and are independent. For convenience, define the random variables $\rg_1:=\rvw^\top \vu$, and $\rg_2:=\rvw^\top \vv$. Let $(s,t)$ be the coordinates in the 2 dimensional space spanned by $\vu$ and $\vv$. Then, note from \eqref{eq:lem2ineq} that we are interested in the behaviour of the hyperplane 
\begin{align}
\rg_1 s +\rg_2 t + \rb=0,\label{eq:line}
\end{align} 
where $\rg_1,\rg_2$ are the coefficients and $\rb$ is the bias.
Note that this is just a line in 2-dimensions. In particular, \eqref{eq:lem2ineq} is still equivalent to the event that this line passes in between $\vx_i$ and $\vx_j$ in this 2-dimensional space.

Looking at the line \eqref{eq:line}, its slope is $-\rg_1/\rg_2$ and its (signed) distance from the origin is $\rb/\sqrt{\rg_1^2+\rg_2^2}$. Next, note that if we consider the two dimensional Gaussian vector $[\rg_2\ \rg_1]^\top \sim \cN(\vZero,\mI_2)$, then thinking in terms of polar coordinates, $\sqrt{\rg_1^2+\rg_2^2}$ is its norm and $-\rg_1/\rg_2$ is its direction. By the spherical symmetry of Gaussians, we know that the norm and its direction are independent. This implies that $\rb/\sqrt{\rg_1^2+\rg_2^2}$ is independent of $-\rg_1/\rg_2$, that is the distance of the line from the origin is independent of its slope. Note that $r:=\sqrt{2}\rb/\sqrt{\rg_1^2+\rg_2^2}$ is the (scaled) ratio of a Gaussian random variable with the square-root of a $\chi^2$-random variable, and hence $r$ has a Student's $t$-distribution with 2 degrees of freedom.

Fix the slope $\theta:=-\rg_1/\rg_2$ of the line \eqref{eq:line}, then we have seen that the signed distance from the origin is the independent random variable $\rr/\sqrt{2}$. Then, for the line \eqref{eq:line} to pass in between $\vx_i$, $\vx_j$, we will need to have that $\frac{\rr}{\sqrt{2}}$ should be between $\left(\rk\frac{z_{i,2}+\theta z_{i,1}}{\sqrt{1+\theta^2}}\right)$ and $\left(\rk\frac{z_{j,2}+\theta z_{j,1}}{\sqrt{1+\theta^2}}\right)$, where $\rk=-1$ or $\rk=1$ depending on the sign of $\theta$. These two exact values do not matter much. What is more important is that since $\|\vx_i\|\leq 1$ and $\|\vx_j\|\leq 1$, we know that these values are less than 1; and that 
\begin{align*}
\left|\left(\rk\frac{z_{i,2}+\theta z_{i,1}}{\sqrt{1+\theta^2}}\right)-\left(\rk\frac{z_{j,2}+\theta z_{j,1}}{\sqrt{1+\theta^2}}\right)\right|&=\frac{|\theta|}{\sqrt{1+\theta^2}}|z_{j,1}-z_{i,1}|\tag*{(Since $\rk\in\{-1,1\}$ and  $z_{i,2}=z_{j,2}$.)}\\
&\geq \delta \frac{|\theta|}{\sqrt{1+\theta^2}}.
\end{align*}

In short, once we fix the slope, $\theta$, the probability that the line passes in between the two points is the same as that of a scaled Student's $t$-variable's value being in an interval of length at least $\delta \frac{|\theta|}{\sqrt{1+\theta^2}}$. Note that the interval lies completely in $[-1,1]$. Because in the interval $[-\sqrt{2},\sqrt{2}]$, a Student's $t$-variable has its p.d.f. lower bounded by a univeral constant $C$, we get that the probability that the line passes in between the two points is at least $2C\sqrt{2}\delta \frac{|\theta|}{\sqrt{1+\theta^2}}$.

To compute a lower bound on the final probability, we need to unfix the slope $\theta$ and integrate the lower bound we computed above, over the distribution of the slope. Define the polar angle $\phi:=\arctan (\theta) = \arctan (-g_1/g_2)$.
Note that due to the spherical symmetry of the Gaussian vector $[\rg_1\ \rg_2]$, we will have that $\phi$ is uniformly distributed in $[0,2\pi]$. Hence, the final probability is lower bounded by
\begin{align*}
\frac{1}{2\pi}\int_0^{2\pi} C\sqrt{2}\delta \frac{|\theta|}{\sqrt{1+\theta^2}} \dPhi = \frac{1}{2\pi}C\sqrt{2}\delta\int_0^\pi  |\sin (\phi)| \dPhi = C'\delta,
\end{align*}
for some universal constant $C'$.

What we have shown above, is that the probability that a hyperplane separates $\vx_i$ and $\vx_j$ is at least $C'\delta$. Hence, the probability that the hyperplane does not separate $\vz_i$ and $\vz_j$, is less than $1-C'\delta$. Since we have $m$ independently sampled hyperplanes, the probability that none of them separate $\vz_i$ and $\vz_j$, is less than $(1-C'\delta)^m$. This is the probability that one of the pairs $\vx_i$ and $\vx_j$ does not get separated. Since there are a total of $\binom{n}{2}$ pairs of points $\vx_i$ and $\vx_j$, then by using the union bound, the probability that there exists one pair of points such that the $m$ hyperplanes does not separate them, is less than $\binom{n}{2}(1-C'\delta)^m$. We want this probability to be less than $\epsilon$, that is,
\begin{align*}
&\binom{n}{2}\left(1-C'\delta\right)^m \leq \epsilon.
\end{align*}
Note that $\binom{n}{2}\leq n^2$ and $\left(1-C'\delta\right)^m\leq e^{-C'\delta m}$. Hence, it is sufficient for the following to hold
\begin{align*}
n^2e^{-C'\delta m} \leq \epsilon.
\end{align*}
This gives that $m\geq \frac{1}{C'\delta}\log \left(\frac{n^2}{\epsilon}\right)$ is sufficient for the lemma to be true.

\end{proof}
\subsubsection{Proof of Lemma~\ref{lem:GVLemma}}\label{subsec:GVLemma}
\begin{proof}
Consider two binary vectors $\vz_i$ and $\vz_j$ that differ in at least one bit. Without loss of generality, assume that the first bit is one of the bits that they differ in. Let $z_{i,k}$ be the $k$-th element of $\vz_i$ and similarly let $z_{j,k}$ be the $k$-th element of $\vz_j$. We can further assume without loss of generality that $z_{i,1}=0$ and $z_{j,1}=1$. Then, for any Bernoulli(0.5) vector $\rvb$, we have
\begin{align*}
\langle \vz_i,\rvb\rangle_{\oplus} &= (0\cdot \rb_1)\oplus (z_{i,2}\cdot \rb_2)\oplus\dots \oplus (z_{i,{d'}}\cdot \rb_{d'}),\\
\langle \vz_j,\rvb\rangle_{\oplus} &= (1\cdot \rb_1)\oplus (z_{j,2}\cdot \rb_2)\oplus\dots \oplus (z_{j,{d'}}\cdot \rb_{d'}),
\end{align*}
where $\rb_k$ is the $k$-th element of $\rvb$. There can be two cases:
\begin{itemize}
\item $(z_{i,2}\cdot \rb_2)\oplus\dots \oplus (z_{i,{d'}}\cdot \rb_{d'})=(z_{j,2}\cdot \rb_2)\oplus\dots \oplus (z_{j,{d'}}\cdot \rb_{d'})$. Say the probability of this is $p$. Note that $\rb_1$ is independent of every other $\rb_i$. With probability $0.5$, $\rb_1=1$. In this case, 
\begin{align*}
\langle \vz_i,\rvb\rangle_{\oplus} &= (0\cdot 1)\oplus (z_{i,2}\cdot \rb_2)\oplus\dots \oplus (z_{i,{d'}}\cdot \rb_{d'})\\
&= 0\oplus (z_{i,2}\cdot \rb_2)\oplus\dots \oplus (z_{i,{d'}}\cdot \rb_{d'})\\
&=  (z_{i,2}\cdot \rb_2)\oplus\dots \oplus (z_{i,{d'}}\cdot \rb_{d'})\\
&=  \lnot (\lnot ((z_{i,2}\cdot \rb_2)\oplus\dots \oplus (z_{i,{d'}}\cdot \rb_{d'})))\tag*{(Applying two NOT operations does not affect the value.)}\\
&= \lnot ( 1 \oplus (z_{i,2}\cdot \rb_2)\oplus\dots \oplus (z_{i,{d'}}\cdot \rb_{d'}))\tag*{(XOR with 1 is equivalent to the NOT operation.)}\\
&= \lnot ( (1\cdot 1) \oplus (z_{i,2}\cdot \rb_2)\oplus\dots \oplus (z_{i,{d'}}\cdot \rb_{d'}))\\
&= \lnot \langle \vz_j,\rvb\rangle_{\oplus}.
\end{align*}
Hence, in this case, with probability at least $0.5$, $\langle \vz_i,\rvb\rangle_{\oplus}\neq \langle \vz_j,\rvb\rangle_{\oplus}$.

\item $(z_{i,2}\cdot \rb_2)\oplus\dots \oplus (z_{i,{d'}}\cdot \rb_{d'})=\lnot ((z_{j,2}\cdot \rb_2)\oplus\dots \oplus (z_{j,{d'}}\cdot \rb_{d'}))$. The probability of this is $1-p$. Note that $\rb_1$ is independent of every other $\rb_i$. With probability $0.5$, $\rb_1=0$. In this case, 
\begin{align*}
\langle \vz_i,\rvb\rangle_{\oplus} &= (0\cdot 0)\oplus (z_{i,2}\cdot \rb_2)\oplus\dots \oplus (z_{i,{d'}}\cdot \rb_{d'})\\
&= 0\oplus (z_{i,2}\cdot \rb_2)\oplus\dots \oplus (z_{i,{d'}}\cdot \rb_{d'})\\
&=  (z_{i,2}\cdot \rb_2)\oplus\dots \oplus (z_{i,{d'}}\cdot \rb_{d'})\\
&=  \lnot((z_{j,2}\cdot \rb_2)\oplus\dots \oplus (z_{j,{d'}}\cdot \rb_{d'}))\\
&=  \lnot((1\cdot 0)\oplus(z_{j,2}\cdot \rb_2)\oplus\dots \oplus (z_{j,{d'}}\cdot \rb_{d'}))\\
&=  \lnot \langle \vz_j,\rvb\rangle_{\oplus}
\end{align*}
Hence, in this case as well, with probability at least $0.5$, $\langle \vz_i,\rvb\rangle_{\oplus}\neq \langle \vz_j,\rvb\rangle_{\oplus}$.
\end{itemize}
Hence overall, with probability at least $p(0.5)+(1-p)(0.5)=0.5$, we have that $\langle \vz_i,\rvb\rangle_{\oplus}\neq \langle \vz_j,\rvb\rangle_{\oplus}$. If we take $m$ i.i.d. Bernoulli(0.5) vectors $\rvb_1,\dots,\rvb_m$, then the probability that 
\begin{align*}
\forall k\in[m], \langle \vz_i,\rvb_k\rangle_{\oplus} = \langle \vz_j,\rvb_k\rangle_{\oplus},
\end{align*}
is less than $(0.5)^{m}$. Equivalently,
\begin{align*}
\Pr({\vz}_i^{(2)}={\vz}_j^{(2)})\leq 2^{-m}.
\end{align*}
To get the bound required in the lemma statement, we simply take a union bound over all the $\binom{n}{2}$ pairs $({\vz}_i^{(2)},{\vz}_j^{(2)})$. Concretely, we want $\Pr(\exists i,j: {\vz}_i^{(2)}={\vz}_j^{(2)})\leq \epsilon$. Hence, we take a union bound:
\begin{align*}
\Pr(\exists i,j: {\vz}_i^{(2)}={\vz}_j^{(2)})&\leq \binom{n}{2} \Pr({\vz}_i^{(2)}={\vz}_j^{(2)})\\
&\leq \binom{n}{2}2^{-m}.
\end{align*}
It can be easily verified that $m\geq 3 \log \frac{n}{\epsilon}$ suffices for $\binom{n}{2}2^{-m}\leq \epsilon$. This completes the proof.
\end{proof}
\subsubsection{Proof of Lemma~\ref{lem:thresh_to_xor}}
\begin{figure}[t]
    \centering
    \includegraphics[width=\textwidth]{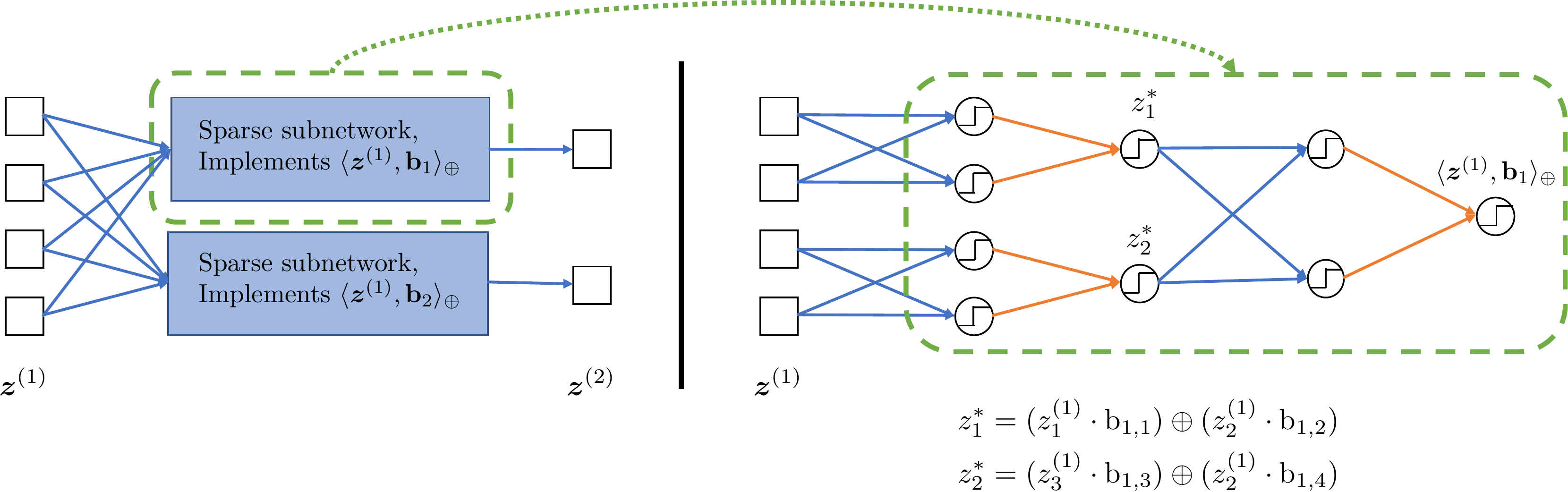}
    \caption{\textbf{Left:} We can compute each of $\langle \vz^{(1)},\rvb_k\rangle$ using subnetworks shown in blue, that we can put in parallel to output $\vz^{(2)}$. \textbf{Right:} A subnetwork that implements $\langle \vz^{(1)},\rvb_1\rangle$. This, in turn can be broken down as $\langle \vz^{(1)},\rvb_1\rangle_\oplus=(z_{1}^{(1)}\cdot \rb_{1,1})\oplus(z_{2}^{(1)}\cdot \rb_{1,2})\oplus\dots\oplus(z_{{d^{(1)}}}^{(1)}\cdot \rb_{1,{d^{(1)}}})$. Then, the figure on the right shows how the network computes $\langle \vz^{(1)},\rvb_1\rangle$ using a binary tree structure (see the orange lines in the figure). In the figure, we take $d^{(1)}=4$. The network first computes $z_1^* = (z_1^{(1)}\cdot \rb_{1,1})\oplus (z_2^{(1)}\cdot \rb_{1,2})$ and $z_2^* = (z_3^{(1)}\cdot \rb_{1,3})\oplus (z_4^{(1)}\cdot \rb_{1,4})$. Then, it computes $z_1^*\oplus z_2^*$, as shown in the figure.}
    \label{fig:XOR_implementation}
\end{figure}
\begin{proof}
We will show that we can compute $\langle \vz_i^{(1)},\rvb_k\rangle_{\oplus}$ with a network of $3(d^{(1)}-1)$ neurons and $9(d^{(1)}-1)$ weights. Then, we can put $m$ such networks in parallel (see Figure\ref{fig:XOR_implementation}, Left) to output
\begin{align*}
{\vz}_i^{(2)}=\begin{bmatrix}
\langle \rvb_1,\vz_i\rangle_{\oplus}\\
\vdots\\
\langle \rvb_m,\vz_i\rangle_{\oplus}
\end{bmatrix}.
\end{align*}

Hence, all we need to do is to prove that we can compute $\langle \vz_i^{(1)},\rvb_k\rangle_{\oplus}$ with a network of $3(d^{(1)}-1)$ neurons and $9(d^{(1)}-1)$ weights. We will do this in a hierarchical fashion as shown in Figure~\ref{fig:XOR_implementation},~Right. We begin by expanding $\langle \vz_i^{(1)},\rvb_k\rangle_\oplus$:
\begin{align*}
\langle \vz_i^{(1)},\rvb_k\rangle_\oplus=(z_{i,1}^{(1)}\cdot \rb_{k,1})\oplus(z_{i,2}^{(1)}\cdot \rb_{k,2})\oplus\dots\oplus(z_{i,{d^{(1)}}}^{(1)}\cdot \rb_{k,{d^{(1)}}}).
\end{align*} 
Note that 
\begin{align*}
(z_{i,1}^{(1)}\cdot \rb_{k,1})\oplus(z_{i,2}^{(1)}\cdot \rb_{k,2})=\sigma(\sigma((z_{i,1}^{(1)}\cdot \rb_{k,1}) -(z_{i,2}^{(1)}\cdot \rb_{k,2})-1)+\sigma(-(z_{i,1}^{(1)}\cdot \rb_{k,1}) +(z_{i,2}^{(1)}\cdot \rb_{k,2})-1))),
\end{align*} 
which uses 3 neurons, each with 2 weights and a bias. 
Similarly, we can compute $(z_{i,3}^{(1)}\cdot \rb_{k,3})\oplus(z_{i,4}^{(1)}\cdot \rb_{k,4})$, $(z_{i,5}^{(1)}\cdot \rb_{k,5})\oplus(z_{i,6}^{(1)}\cdot \rb_{k,6})$, \dots, $(z_{i,d^{(1)}-1}^{(1)}\cdot \rb_{k,d^{(1)}-1})\oplus(z_{i,d^{(1)}}^{(1)}\cdot \rb_{k,d^{(1)}})$. 
Once we have computed these, we can compute XOR's of four terms similarly, to get terms like $(z_{i,1}^{(1)}\cdot \rb_{k,1})\oplus(z_{i,2}^{(1)}\cdot \rb_{k,2})\oplus $.
We keep applying this hierarchically in a binary tree form to get $\langle \vz_i,\rvb_k\rangle_\oplus(z_{i,3}^{(1)}\cdot \rb_{k,3})\oplus(z_{i,4}^{(1)}\cdot \rb_{k,4})$ (see Figure~\ref{fig:XOR_implementation}, Right). The tree will consist of $d^{(1)}-1$ XOR operations, each represented by a node. For each of these operations, we saw above that we need 3 neurons, each with 2 weights and a bias. This gives that we need $3(d^{(1)}-1)$ neurons and $9(d^{(1)}-1)$ weights and biases to compute $\langle \vz_i^{(1)},\rvb_k\rangle_\oplus$.

\end{proof}
\subsubsection{Proof of Lemma~\ref{lem:prefix}}
\begin{proof}
Consider a binary tree of depth $d^{(2)}+1$, were the left child edge of each node is labeled 0, and the right one is labeled 1. There will be $2^{d^{(2)}}$ leaves, each representing one unique binary vector of length $d^{(2)}$. The vector associated with a leaf would be the same as the sequence of edges of the path from the root to that leaf. Hence, $\{{\vz}_1^{(2)},\dots,{\vz}_n^{(2)}\}$ will form a subset of the $2^{d^{(2)}}$ leaves. Remove all other leaves from the tree, so that we only have $\{{\vz}_1^{(2)},\dots,{\vz}_n^{(2)}\}$ as the leaves in our tree. Hence, now our tree only has $n$ leaves. Also remove all the nodes that are not the ancestor of any leaves.

The way we select subsets is simple: Each node is a potential subset. Each node represents the subset of $\{{\vz}_1^{(2)},\dots,{\vz}_n^{(2)}\}$ that belong in the subtree at that node. Each node is also associated with a unique prefix: the sequence of edges of the path from the root to that node. Denote by $\cS$ the set of nodes (or equivalently subsets) that we select. We include a node in $\cS$ , if 
\begin{enumerate}
\item the number of leaves in the subtree at that node is at most $\sqrt{n}$, 
\item none of its ancestor nodes satisfy condition 1.
\end{enumerate}

Condition 1 will ensure that the subsets formed in this way will have size at most $\sqrt{n}$. Condition 2 above ensures that the subsets do not overlap and consequently the prefixes of the selected nodes are unique. 
Hence, all that is left to do is prove that $|\cS|=\cO(\sqrt{n}\log n)$. 

Define the height of a node as $(d^{(2)}-\text{its depth})$. Thus, the root node has height $d^{(2)}$ and the leaves have height $0$. We will prove that for any node at height $h$, if any of its descendant nodes is chosen, then 
\begin{align}
\frac{\#\text{ leaves in the subtree at that node}}{\#\text{ descendant nodes included in $\cS$}} \geq \frac{\sqrt{n}}{h}.\label{eq:ratio}
\end{align}
At root node, $h=d^{(2)}=c\log n$, and $\#$ leaves $=n$. Then, rearranging the inequality \eqref{eq:ratio}, we get that at the root node the number of descendant nodes included in $\cS$ are less than $\sqrt{n}(c\log n)$. Thus, the total number of subsets is $\cO(\sqrt{n}\log n)$. Hence, all we need to do now, is to prove \eqref{eq:ratio}. For this, we will use induction on $h$.

The smallest $h$ for which the inequality is applicable, will be $h=\lceil \log_2 \sqrt{n} \rceil+$. This is because the first height at which a node can be selected is $\lceil \log_2 \sqrt{n} \rceil$. Note that no descendants of a node at height $h=\lceil \log_2 \sqrt{n} \rceil+1$, other than its children, can be included in $\cS$. Further, since one of its children is included in $\cS$, this node itself is not included in $\cS$. Hence, the number of leaves in the subtree at this node is at least $\sqrt{n}+1$. Hence for this node,
\begin{align*}
\frac{\#\text{ leaves in the subtree formed at that node}}{\#\text{ descendant nodes in $\cS$}} \geq \frac{\sqrt{n}+1}{2}\geq \frac{\sqrt{n}}{h}. 
\end{align*}
Thus, the base case is proved. We move on to the inductive case now: $h>\lceil \log_2 \sqrt{n} \rceil+1$. There can be the following cases for a node $s$ at height $h$.
\begin{itemize}
\item $s$ has only one child. This child cannot be in $\cS$, since Condition 2 would be violated. Thus, it can only be the case that this child has further descendants that are in $\cS$. Then, by the induction hypothesis, \eqref{eq:ratio} is satisfied for the child, and hence it is also satisfied for $s$.
\item Both of its children have descendants that are in $\cS$. Let $l_{\text{l}}$ be the $(\#$ leaves in the subtree formed at the left child of $s)$ and $d_\text{l}$ be the $(\#\text{ descendant nodes of the left child in $\cS$})$. Similarly, we define $l_\text{r}$ and $d_\text{r}$ for the right child. Then, we know by the induction hypothesis that 
\begin{align*}
\frac{l_{\text{l}}}{d_\text{l}}\geq \frac{\sqrt{n}}{h-1},\frac{l_{\text{r}}}{d_\text{r}}\geq \frac{\sqrt{n}}{h-1}.
\end{align*}
Hence, for the node $s$,
\begin{align*}
\frac{\#\text{ leaves in the subtree formed at $s$}}{\#\text{ descendant nodes in $\cS$}}&=\frac{l_{\text{l}}+l_{\text{r}}}{d_\text{l}+d_\text{r}}\\
&\geq \frac{\frac{\sqrt{n}}{h-1}(d_{\text{l}}+d_{\text{r}})}{d_\text{l}+d_\text{r}}\\
&\geq \frac{\sqrt{n}}{h}.
\end{align*}
\item One of its children is included in $\cS$ and the other node has at least one of its descendants included in $\cS$. Let the left child be the one that is included in $\cS$. Let $l_{\text{r}}$ be the $(\#$ leaves in the subtree formed at the right child of $s)$ and $d_\text{r}$ be the $(\#\text{ descendant nodes of the right child included in }\cS)$. Then, by inductive hypothesis,
\begin{align*}
\frac{l_{\text{r}}}{d_\text{r}}\geq \frac{\sqrt{n}}{h-1}.
\end{align*}
We also know that since the right child wasn't included in $\cS$, $l_{\text{r}}>\sqrt{n}$. Note that the number of leaves under $s$ is at least $l_r$ and the number of its descendants that are in $\cS$ is $d_\text{r}+1$, including the left child. Then,
\begin{align*}
\frac{\#\text{ leaves in the subtree formed at $s$}}{\#\text{ descendant nodes in $\cS$}}&\geq \frac{l_{\text{r}}}{1+d_\text{r}}\\
&= \frac{1}{\frac{1}{l_{\text{r}}}+\frac{d_\text{r}}{l_{\text{r}}}}\\
&\geq \frac{1}{\frac{1}{l_{\text{r}}}+\frac{h-1}{\sqrt{n}}}\\
&= \frac{\sqrt{n}}{\frac{\sqrt{n}}{l_{\text{r}}}+h-1}\\
&> \frac{\sqrt{n}}{{h}}.
\end{align*}
\item Both the children of the node are in $\cS$. In this case, since the node itself wasn't chosen, the number of leaves in the subtree is at least $\sqrt{n}+1$, and the number of descendant nodes in $\cS$ is 2. Hence,
\begin{align*}
\frac{\#\text{ leaves in the subtree formed by that node}}{\#\text{ descendant nodes in $\cS$}} \geq \frac{\sqrt{n}+1}{2}\geq \frac{\sqrt{n}}{h}. 
\end{align*}
\end{itemize}
This proves the inductive case and completes the proof of lemma. 
\end{proof}

\section{Proof of Theorem~\ref{thm:lowerBound} and Theorem~\ref{thm:hyperplaneSeparation}}

\begin{proof}
We want to construct an example where there are $n$ points on the unit sphere $\cS^{d-1}$, which are separated by a distance at least $\delta$, such that the minimum number of hyperplanes needed for separating each pair is ${\varOmega}(\log(n)/\delta)$ and for separating pairs with opposite labels is ${\varOmega}(1/\delta)$. 

Before we start the proof, let us define some notation. We denote by $\cS^{d-1}:=\{x:\|x\|=1\}$, the unit sphere in $d$-dimensions centered at the origin. We can also have lower dimensional spheres in the same space. 
For example, we can have circles in a 3-dimensional space. We denote by $\cS^{d-2}_{\vc}$, the $d-1$ dimensional unit sphere, centered at $\vc$ and which lies on the hyperplane orthogonal to the vector $\vc$. For example, in 3-dimensions, let $\vc=(1,1,1)$. 
Let $\cH_\vc$ be the hyperplane that passes through $\vc$ and is orthogonal to the vector connecting the origin to $\vc$. On this hyperplane, we can draw a unit circle, centered at $(1,1,1)$, which is what we denote by $\cS^{2}_{(1,1,1)}$ in our notation. Further, $r\cS^{d-1}$ and $r\cS^{d-2}_\vc$ will denote the corresponding hyperspheres, but with radius $r$ instead of 1. We denote the cluster centered at $\vc$ by $\cQ_\vc$. We will use $C_1,C_2,\dots$ to denote universal constants.

Now, we are ready to start the construction. We will create $\sqrt{n}$ clusters of points on the sphere, each of size $\sqrt{n}$. 

First, the centers of the $\sqrt{n}$ clusters are sampled uniformly from $(\sqrt{1-\delta})\cS^{d-1}$.
Then, in each cluster, $\sqrt{n}$ points are sampled i.i.d. from $\sqrt{\delta}\cS^{d-2}_{\vc}$, where $\vc$ is the center of the cluster.
This way, each point has norm $\sqrt{(\sqrt{1-\delta})^2+(\sqrt{\delta})^2}=1$, that is, each point indeed lies on $\cS^d$. Of the $\sqrt{n}$ points in each cluster, we label half of them as $0$ and the other half as $1$.

Then, roughly speaking, the summary of the proof is as follows:
\begin{itemize}
	\item Any hyperplane can only `effectively' separate points within a cluster if it passes close to the center of the cluster. The distance would roughly need to be $\widetilde{O}(\sqrt{\delta} /\sqrt{d})$.
	\item A hyperplane can be $\widetilde{O}(\sqrt{\delta}/\sqrt{d})$ close to the centers of only $\widetilde{O}(\sqrt{\delta})$ fraction of the clusters.
	\item Any cluster needs at least
	\begin{itemize}
	    \item $\varOmega(\log_2 \sqrt{n})=\varOmega(\log n)$ hyperplanes to separate each pair of points within the cluster (for Theorem~\ref{thm:hyperplaneSeparation}) or
	    \item at least 1 hyperplane to separate all the opposite label pairs (for Theorem~\ref{thm:lowerBound}).
	\end{itemize}
	\item Finally, the core inequality that we use is that  

\begin{align}
    (\#\text{ hyperplanes needed}) \times (\#\text{ clusters a hyperplane can `effectively' separate}) \nonumber\\
    \ \ \geq (\#\text{ clusters})\times (\#\text{ hyperplanes needed per cluster}).\label{eq:core}
\end{align}

	Substituting the bounds we got above into this inequality, we get that we need at least $\varOmega(\log(n)/\sqrt{\delta})$ hyperplanes.
\end{itemize}

Now we describe the construction in detail.
Note that we only need to prove the existence of \emph{one} such construction. 
Thus, we will use the probabilistic method. 

We define four events:
\begin{enumerate}[label=$\cE_{\arabic*}:$]
\item Two points from different clusters have distance less than $\delta$.
\item The points within a cluster have distance less than $\delta$.
\item A hyperplane at distance $\varOmega( \sqrt{\frac{\delta\log t}{d}})$ from the center of a cluster separates $\varOmega(n/t)$ pairs of points in a cluster, for $t\geq 4$.
\item A hyperplane passes within distance $O( \sqrt{\frac{\delta\log t }{d}})$ of the centers of $\varOmega(\sqrt{\delta n\log t})$ clusters.
\end{enumerate}
We will prove that the probability of each of these events is at most $1/5$. Hence, a union bound shows that there exists an event where none of these events happen. We will prove that such an event gives the required lower bounds.

\paragraph*{Step 1: Bound the probability of $\cE_1$.}

Let $\vc_1,\dots,\vc_{\sqrt{n}}$ be $\sqrt{n}$ cluster centers sampled uniformly from $(\sqrt{1-\delta})\cS^{d-1}$, representing the centers of the clusters. Then there exists a universal constant $C$, such that, as long as $n<\left(\frac{C}{\sqrt{\delta}}\right)^{d-1}$, we show that the event that two points from different clusters have distance less than $\delta$ will have probability less than $1/5$.

To see how this is true, consider any two clusters, with centers $\vc_i$ and $\vc_j$. Then, as long as $\|\vc_i-\vc_j\|\geq 2\sqrt{\delta}+\delta$, we get by triangle inequality that two points from these two clusters are at least $\delta$ distance apart. Since $\sqrt{\delta}\geq \delta$, we know that if we ensure that all cluster centers are at least $3\sqrt{\delta}$ apart, then the above still holds.

Hence, we need to show that if we sample $\sqrt{n}$ points (cluster centers) on a sphere of radius $\sqrt{1-\delta}$, and $n<\frac{1}{5}\left(\frac{C}{\sqrt{\delta}}\right)^{d-1}$, then they are at least $3\sqrt{\delta}$ apart with probability at least $4/5$. For this, we need to show that the probability of any pair of points $\vc_i$ and $\vc_j$ being within $3\sqrt{\delta}$ distance is less than $\left(\frac{\sqrt{\delta}}{C}\right)^{d-1}$. 
Taking a union bound over all the $\binom{\sqrt{n}}{2}$ pairs and substituting $n<\left(\frac{C}{\sqrt{\delta}}\right)^{d-1}$ will give that the probability of any such pair existing is less than $1/5$.

Thus, all we need to do is to prove that the probability of any pair of points being within $3\sqrt{\delta}$ distance is less than $\left(\frac{\sqrt{\delta}}{C}\right)^{d-1}$. We prove it below.

Fix the point $\vc_i$. Then, the probability of any other cluster center $\vc_{j}$ being within distance $3\sqrt{\delta}$ of $\vc_i$ is the same as the ratio of the surface area of the cap of radius $3\sqrt{\delta}$ to the surface area of $\sqrt{1-\delta}\cS^{d-1}$. Consider packing the surface of $\sqrt{1-\delta}\cS^{d-1}$ with caps of radius $3\sqrt{\delta}$, and let $\cA_{3\sqrt{\delta}}$ denote such a cap. Let the number of caps required for the packing be
$\cP'$. Then, 
\begin{align*}
    &\text{Area}(3\sqrt{\delta} \text{ cap})\times \cP'\leq \text{Area}(\sqrt{1-\delta}\cS^{d-1})\\
    \implies & \frac{\text{Area}(3\sqrt{\delta} \text{ cap})}{\text{Area}(\sqrt{1-\delta}\cS^{d-1})}\leq\frac{1}{ \cP'}.
\end{align*}

Further, note that by rescaling $|\cP'|$ is equal to $|\cP_{3\sqrt{\delta}/\sqrt{1-\delta}}|$, where $\cP_{\delta'}$ is the $\delta'$ packing of $\cS^{d-1}$. Finally, we use the following lemma (proved later in the appendix) to conclude that the probability of any pair of points being within $3\sqrt{\delta}$ distance is less than $\left(\frac{\sqrt{\delta}}{C}\right)^{d-1}$, for some universal constant $C$.
\begin{lemma}\label{lem:sphere_packing}
Let $\cC_\delta$ and $\cP_\delta$ denote the covering and packing numbers of $\cS^{d-1}$, respectively. Then,
\begin{align*}
    \left(\frac{1}{4\delta}\right)^{d-1} \leq \cC_\delta \leq \cP_\delta\leq \left(\frac{2}{\delta}\right)^d.
\end{align*}
\end{lemma}

\paragraph*{Step 2: Bound the probability of $\cE_2$.}
Consider any one cluster center $\vc$.
 Then, the sphere $\sqrt{\delta}\cS_c^{d-2}$ is just a $d-2$ dimensional sphere. We create this cluster by sampling points uniformly on $\cS_c^{d-2}$. This is similar to what we did in previous step and hence we can apply exactly the same steps as above to get that as long as $n<\left(\frac{C}{\sqrt{\delta}}\right)^{d-2}$, we have that the event that two points within some cluster have distance less than $\delta$, will have probability less than $1/5$. We denote this cluster by  $\cQ_\vc$. Then, $\cQ_\vc$ lies on the sphere $\sqrt{\delta}\cS_\vc^{d-2}$, which in turn lies on the hyperplane $\cH_\vc$.

We need to create such clusters around each of the other cluster centers.
A trick that we employ is that we use the same cluster $\cQ_\vc$ and `paste' it at every other cluster center, after rotating appropriately to make the cluster lie on the hyperplane orthogonal to the corresponding cluster center. This is helpful because we would not have to take a union bound of the probability calculation done above for every cluster. We formally describe the procedure below.

Let $\vc'$ be another cluster center. Then, there will be a corresponding sphere $\sqrt{\delta}\cS_{\vc'}^{d-2}$, which lies on the hyperplane $\cH_{\vc'}$. Let $T_{c'}$ be any linear transformation such that $T(\sqrt{\delta}\cS_{\vc}^{d-2})=\sqrt{\delta}\cS_{\vc'}^{d-2}$. Then, we create the cluster $\cQ_{\vc'}$ for the cluster center $\vc'$ as $\cQ_{\vc'}=T(\cQ_{\vc})$.

\paragraph*{Step 3: Bound the probability of $\cE_3$.}
Any hyperplane $\cH_2$ (distinct from $\cH_\vc$) that intersects $\cH_\vc$ creates a $d-1$ dimensional hyperplane, say $\cH_2'$. We want to show that if $\cH_2$ lies at distance $\varOmega( \sqrt{\frac{\delta\log t }{d}})$ from a cluster center $\vc$, then the probability that it separates $\geq n/ t$ pairs of points in the cluster is less than $1/5$. 
We will actually show that if $\cH_2'$ lies at distance $\varOmega( \sqrt{\frac{\delta\log t }{d}})$ from $\vc$, then the probability that it separates $\varOmega(n/ t)$ pairs of points in the cluster is less than $1/5$, which is a stronger statement because the distance of $\vc$ from $\cH_2$ is smaller than the distance of $\vc$ from $\cH_2'$. We do not consider hyperplanes parallel to $\cH_\vc$ because such hyperplanes do not separate any points in $\cQ_\vc$. We will work in the $d-1$ dimensional subspace spanned by $\cH_\vc$, with $\vc$ as the origin.

Take any one direction in this space.
The probability measure of $\sqrt{\delta}\cS_\vc^{d-2}$ at distance $\geq 2\sqrt{ \frac{\delta\log t}{d-1}}$ in this direction is at most $e^{-2 \log t}$ \citep[Lemma 2.2]{ball1997elementary}. 
Thus, the probability that a point in the cluster lies at a distance more than $2\sqrt{ \frac{\delta\log t}{d-1}}$ is less than $t^{-2}$.
We can now apply Binomial tail probability bounds to get a probability bound on the event $$E: \left\{\text{more than $\frac{1}{t}$ fraction of the $\sqrt{n}$ points lie at a distance more than $2\sqrt{ \frac{\delta\log t}{d-1}}$ in one fixed direction}\right\}.$$ Let $D(\cdot\|\cdot)$ denote the KL-divergence. Then, the Chernoff bound on the tail probability of a Binomial random variable \citep{arratia1989tutorial} gives the following bound
\begin{align*}
\Pr(E)&\leq e^{-\sqrt{n}D(\frac{1}{t}\|\frac{1}{t^2})}\\
&= \exp\left(-\sqrt{n}\left(\frac{1}{t}\log t + \left(1-\frac{1}{t}\right)\log \frac{1-\frac{1}{t}}{1-\frac{1}{t^2}} \right)\right)\\
&= \exp \left(-\sqrt{n}\left(\frac{1}{t}\log t + \left(1-\frac{1}{t}\right)\log \frac{t}{1+t}\right )\right)\\
&= \exp \left(-\sqrt{n}\left( \log t -\log (1+t) +  \frac{1}{t}\log (1+t+t^2)\right )\right)\\
&= \exp \left(-\sqrt{n}\left( \log\left( \frac{t}{1+t}\right)  +  \frac{1}{t}\log (1+t)\right )\right)\\
&= \exp \left(-\sqrt{n}\left( \log\left( 1-\frac{1}{1+t}\right)  +  \frac{1}{t}\log (1+t)\right )\right)\\
&\leq \exp \left(-\sqrt{n}\left( -\frac{1}{2(1+t)}  +  \frac{1}{t}\log (1+t)\right )\right)\tag*{(Since $\frac{1}{1+t}\leq 1$)}\\
&\leq \exp \left(\frac{-\sqrt{n}}{2t}\log (1+t)\right)\tag*{(Since $t\geq 2$)}\\
&\leq \frac{1}{t^{\sqrt{n}/2t}}.
\end{align*}

Recall that the above result is just for one fixed direction, whereas we want to ensure that no more than $1/t$ fraction of points lie at distance $\varOmega(\sqrt{\delta \log t/(d-1)})$ in \emph{any} direction. For this, we need an $\epsilon$-net argument\footnote{An $\epsilon$-net $\cE$ of a set $\cB$ is a set of points from $\cB$ such that any point in $\cB$ is at distance at most $\epsilon$ from some point in $\cE$.} over all the directions. Note that a direction is essentially just a point on the unit sphere and hence, we need to make an $\epsilon$-net on $\cS^{d-2}$. We claim that for $\epsilon=\sqrt{\frac{\log t}{d-1}}$, if we prove that no more than $1/t$ fraction of points lie at distance $2\sqrt{\delta \log t/(d-1)}$ in any direction in the $\epsilon$-net, then no more than $1/t$ fraction of points lie at distance $4\sqrt{\delta \log t/(d-1)}$ in any direction. We will prove this claim shortly. Recall that an $\epsilon$-net is the same as a covering and hence we can use Lemma~\ref{lem:sphere_packing} to get an upper bound on the size of the $\epsilon$-net (Note that the radius of the sphere here is $\sqrt{\delta}$, so we need to rescale appropriately). Taking a union bound over the $\epsilon$-net gives the probability
\begin{align}
p:=\left(4\sqrt{\frac{d-1}{\log t}}\right)^{d-1}\frac{1}{t^{\sqrt{n}/2t}}.\label{eq:prob_sep}
\end{align}

Thus, any hyperplane that lies at distance more than $4\sqrt{\frac{\delta \log t}{d-1}}$ from the center of a cluster has less than $\sqrt{n}/t$ points on one side of it, with probability at least $1-p$. Note that if a hyperplane has less than $\sqrt{n}/t$ points on one side, then it can only separate $\leq \frac{\sqrt{n}}{t}(\sqrt{n}-\frac{\sqrt{n}}{t})\leq \frac{n}{t}$ pairs of points, which is what we want to prove in this step. 
Hence, we want to show that $p<1/5$.
From \eqref{eq:prob_sep}, we can see that this will be true as long as $\sqrt{n}\geq C_3 dt\log d$, where $C_3$ is a large enough universal constant.

To finish this step, we need to prove that an $\epsilon$-net would ensure that in every direction, no more than $1/t$ fraction of points lie beyond $4\sqrt{ \frac{\delta\log t}{d-1}}$. For any direction (or equivalently, a unit vector) $\vu$, there is a direction $\vu_e$ in our epsilon net such that $\|\vu-\vu_e\|\leq 2\sqrt{\frac{\log t}{d-1}}$. Consider the region of $\sqrt{\delta}\cS_\vc^{d-2}$ consisting of points that lie at distance $4\sqrt{ \frac{\delta\log t}{d-1}}$ from the origin (\ie{} $\vc$) in the direction of $\vu$. Let $\vx$ be any point in the region. Then,
\begin{align*}
    \langle \vx, \vu_e\rangle &=\langle \vx, \vu\rangle + \langle \vx, \vu_e-\vu\rangle\\
     &\geq 4\sqrt{ \frac{\delta\log t}{d-1}} + \langle \vx, \vu_e-\vu\rangle\\
     &\geq 4\sqrt{ \frac{\delta\log t}{d-1}} - \|\vx\|\| \vu_e-\vu\|\\
     &\geq 4\sqrt{ \frac{\delta\log t}{d-1}} - 2\sqrt{ \frac{\delta\log t}{d-1}}\\
     &= 2\sqrt{ \frac{\delta\log t}{d-1}}.
\end{align*}
Hence, any point in such a region lies at distance $\geq 2\sqrt{ \frac{\delta\log t}{d-1}}$ in the direction $\vu_e$.

\paragraph*{Step 4: Bound the probability of $\cE_4$.}
This is similar to the previous step. We need to show that there are only $\cO(\sqrt{\delta n\log t})$ cluster centers within distance $\cO(\sqrt{\delta \log t / d})$ of any hyperplane.

Take any fixed hyperplane. Then, the probability measure of the sphere $\sqrt{1-{\delta}}\cS^{d-1}$ within distance $12\sqrt{\delta \log t}/\sqrt{d}$ of the hyperplane is less than $C_5\sqrt{\delta\log t}$, for some universal constant $C_5$. Using this, we get that the probability that any cluster center lies within distance  $12\sqrt{\delta \log t}/\sqrt{d}$ of the hyperplane is less than $C_5\sqrt{\delta\log t}$. Similar to the previous step, we can now use Chernoff bounds to get the probability that more than $2C_5\sqrt{\delta \log t}$ fraction of cluster centers lie at a distance less than $12\sqrt{\delta\log (t) / d}$ is less than $e^{-C_5 \sqrt{n\delta \log t}}$.

Recall that this is for a fixed hyperplane, whereas we want to give a probability bound for all hyperplanes. Hence, similar to the previous step, we take a union bound over an $\epsilon$-net of hyperplanes. We claim that if $\epsilon=4\sqrt{\delta \log (t)/d}$, then the $\epsilon$-net argument will ensure that for any hyperplane, the probability that more than $2C_5\sqrt{\delta \log t}$ fraction of cluster centers lie at a distance less than $4\sqrt{\delta\log (t) / d}$ is less than $e^{-C_5 \sqrt{n\delta \log t}}$. We will prove this shortly. We will see that the size of the net will be $\frac{1}{\epsilon}$ times the the covering number of $\cS^{d-1}$.
Taking a union bound over the $\epsilon$-net gives the probability
\begin{align*}
p':=\frac{1}{\epsilon}\left(C_6\sqrt{\frac{d}{\delta\log t}}\right)^{d-1}e^{-C_5 \sqrt{n\delta \log t}}.
\end{align*}
$p'$ will be less than $1/5$ when $\sqrt{n} > C_7 \frac{d}{\sqrt{\delta}}\log \frac{d}{\delta}$.

Next, we prove that if we create an $\epsilon$-net of hyperplanes, then that would indeed be sufficient to ensure that for any hyperplane, the probability that more than $2C_5\sqrt{\delta \log t}$ fraction of cluster centers lie at a distance less than $4\sqrt{\delta\log (t) / d}$ is less than $e^{-C_5 \sqrt{n\delta \log t}}$. A hyperplane can be represented by $h(\vx)=\vu^\top\vx + b $, where $\vu$ is a unit vector. Similar to previous step, create an $\epsilon$-net over the unit vectors $\vu$. We also create an $\epsilon$-net over $b$. However, since we are only interested in hyperplanes that intersect the sphere, we restrict $b\in [0,1]$, and hence the epsilon net over $b$ would be of size $1/\epsilon$. We know that there is a hyperplane $h_e(\vx)=\vw_e^\top \vx +b_e$ in our $\epsilon$-net such that $\|\vw_e-\vw\|\leq \epsilon$ and $|b-b_e|\leq \epsilon$. Consider the region of $\sqrt{1-\delta}\cS^{d-1}$ within distance $4\sqrt{\delta\log (t) / d}$ of $h$. Hence, for any point $\vx$ in the region,
\begin{align*}
    \vw_e^\top \vx +b_e &= (\vw^\top \vx +b)+ (\vw_e-\vw)^\top\vx + (b_e-b)\\
    &\leq  (\vw^\top \vx +b)+ (\vw_e-\vw)^\top\vx + \epsilon\\
     &\leq  (\vw^\top \vx +b)+ \|\vw_e-\vw\|\|\vx\| + \epsilon\\
     &\leq  (\vw^\top \vx +b)+ \epsilon + \epsilon\\
     &\leq  12\sqrt{\delta\log (t) / d}.
\end{align*}
Hence, any such point lies within distance $12\sqrt{\delta\log (t) / d}$.

\textbf{Step 5: Using the core inequality.}
We showed in the steps above that with probability at least $1/5$, none of the events $\cE_1,\cE_1,\cE_1,\cE_4$ happen. We will work in such an event, \ie{} the event when none of $\cE_1,\cE_1,\cE_1,\cE_4$ happen.

Let $h$ denote the minimum number of hyperplanes needed to separate the dataset that we have constructed. We set $t=8h$. Then, we showed in Step 3 that any hyperplane at distance more than $4\sqrt{\delta \log (t)/(d-1)}$ from a cluster center can only separate $n/t=n/8h$ pairs of points, whereas there are a total of $n/4$ pairs of opposite label points in each cluster. Hence, we need at least one hyperplane to pass within distance $4\sqrt{\delta \log (t)/(d-1)}$ of the cluster center, otherwise there would be at least $n/8$ pairs of opposite label points that would not be separated. On the other hand, we showed in Step 4 that a hyperplane can pass within distance $4\sqrt{\delta \log (t)/(d-1)}$ of only $2C_5\sqrt{\delta \log t}$ fraction of the cluster centers. Now, we are ready to use the core inequality \eqref{eq:core} to get that 
\begin{align*}
    h\geq \frac{\sqrt{n}\times 1}{2C_5\sqrt{\delta \log t}\sqrt{n}}.
\end{align*}
This gives that $h=\varOmega(1/(\sqrt{\delta}\log (1/\delta)))$.

For the case when we need to separate every pair of points (Theorem \ref{thm:hyperplaneSeparation}), the only difference in the inequality above is that we would need at least $\log_2 ({n}-\frac{n}{8})$ hyperplanes per cluster (instead of 1) that need to pass within distance $4\sqrt{\delta \log (t)/(d-1)}$ of the cluster center. This is because separating $m$ points needs $\log_2 m$ hyperplanes. Substituting this in the inequality \eqref{eq:core} gives the bound in Theorem~\ref{thm:hyperplaneSeparation}.
\end{proof}
\subsection{Proof of Lemma~\ref{lem:sphere_packing}}
\begin{proof}
Let $\cP_\delta$ be a $\delta$ packing of the unit sphere $\cS^{d-1}$. Let $\cB^{d}$ denote the unit ball, \ie{} the sphere $\cS^{d-1}$ along with its interior. Then, if we draw balls of radius $\delta$ around each point in the packing, they will not intersect and all of them will be completely contained inside the ball of radius $1+\delta$. Thus, we get the following 
\begin{align*}
    |\cP_\delta|\times \text{Vol}(\delta \cB^{d})&\leq \text{Vol}((1+\delta) \cB^{d})\\
    \implies |\cP_\delta| &\leq \frac{\text{Vol}((1+\delta) \cB^{d})}{\text{Vol}(\delta \cB^{d})}\\
    &= \left(\frac{1+\delta}{\delta}\right)^d\\
    &\leq \left(\frac{2}{\delta}\right)^d.
\end{align*}

Next, let $\cC_\delta$ be a $\delta$-covering of the sphere $\cS^{d-1}$. Any point on the sphere $(1-\delta)\cS^{d-1}$ is at distance $\delta$ of some point on $\cS^{d-1}$. That point is, in turn, at distance within $\delta$ of some point in $\cC_\delta$. Hence, by the triangle inequality, every point on on the sphere $(1-\delta)\cS^{d-1}$ is within distance $2\delta$ of some point in $\cC_\delta$. Similarly, it can be shown that every point on on the sphere $(1+\delta)\cS^{d-1}$ is within distance $2\delta$ of some point in $\cC_\delta$. Draw balls of radius $2\delta$ around each point in the covering $\cC_\delta$. Then, these balls cover all the points within distance $\delta$ of the sphere $\cS^{d-1}$. Thus, we get the following
\begin{align*}
    |\cC_\delta|\times \text{Vol}(2\delta \cB^{d})&\geq \text{Vol}(((1+\delta) \cB^{d})\setminus((1-\delta) \cB^{d})),\\
    \text{which implies} |\cC_\delta| &\geq \frac{\text{Vol}((1+\delta) \cB^{d})-\text{Vol}((1-\delta) \cB^{d})}{\text{Vol}(2\delta \cB^{d})}\\
    &= \frac{(1+\delta)^d-(1-\delta)^d}{(2\delta)^d}\\
    &\geq \frac{(1+\delta)^d-1}{(2\delta)^d}\\ 
    &\geq \frac{1+d\delta-1}{(2\delta)^d}\\
    &= \frac{d\delta}{(2\delta)^d}\\
    &\geq \frac{1}{(4\delta)^(d-1)}.
\end{align*}
Combining with the following inequality \cite[Lemma 4.2.8]{vershynin2018high}:
\begin{align*}
    \cC_\delta \leq \cP_\delta,
\end{align*}
we get
\begin{align*}
    \left(\frac{1}{4\delta}\right)^{d-1} \leq \cC_\delta \leq \cP_\delta\leq \left(\frac{2}{\delta}\right)^d.
\end{align*}
\end{proof}

\section{Proofs from Section~\ref{sec:info_theory}}
\subsection{Proof of Theorem~\ref{thm:info_theoretic_lower_bound}}
\begin{table}[H]
\centering
\begin{tabular}{|l|l|l|l|l|l|}
\hline
$\cH \downarrow, \cP \rightarrow$ & $p_1$ & $p_2$ & $p_3$ & \dots & $p_{|\cP|}$ \\ \hline
$h_1$                                                                                        & 1          & 0    & 0    &                       & 0                           \\\hline
$h_2$                                                                                        & 1          & 1    & 1    &                       & 1                           \\\hline
$h_3$                                                                                        & 0          & 1    & 0    &                       & 1                           \\\hline
\vdots                                                                       &            &      &      &                       &                             \\\hline
$h_{|\cH|}$                                                                 & 0          & 1    & 0    &                       & 1            \\\hline              
\end{tabular}%
\vspace{2em}
\caption{(Lower bound) Here, if there are $r$ rows and $c$ columns, and $2^r<c$, then there will be two columns with the same pattern of 0's and 1's. Noting that $r=|\cH|$ and $c=|\cP|$ gives us the lower bound.}
\label{tab:1}
\end{table}
\begin{proof}
Let $\cA$ output models from the hypothesis set $\cH$. Then, we will essentially try to lower bound $\log_2 |\cH|$, since we will need at least $\log_2 |\cH|$ bits to represent the models.

The first lower bound is easy to see: For a fixed set of $n$ points, there can be $2^n$ possible assignments of labels. Hence we need at least $2^n$ hypothesis in our hypothesis class. Hence we need at least $n$ bits to memorize our dataset. 

A lower bound with dependence on $\delta$ is also not difficult to get. Here is how we create our dataset: Create a $\delta$-packing $\cP_{\delta}$ of the set $\cS$. Choose any $n$ points from this packing. In fact, we will only concentrate on the first and the second point in our dataset. Assume that we have a hypothesis set $\cH$ such that this can memorize these two points. This means that no matter which two points we choose from the packing and what labels we assign them, there will be a hypothesis in $\cH$ that gives the chosen points the assigned labels. Now, if $2^{|\cH|}< |\cP_{\delta}|$, then there exist at least one pair of points in the packing $\cP_\delta$ such that every hypothesis assigns the two points the same labels. 
To see how, consider Table \ref{tab:1}. There can be at most $2^{|\cH|}$ unique columns in that table.
Hence, if $2^{|\cH|}< |\cP_{\delta}|$, then there would exist two points which have the same columns.
We choose those two points and give them opposite labels, then this would contradict memorization, since all the hypotheses in $\cH$ give them both the same label.
This proves that $2^{|\cH|}\geq |\cP_{\delta}|$, or $\log_2 |\cH|\geq \log_2 \log_2 |\cP_{\delta}|$. Combined with the first lower bound, we get that we need at least $\max(n,\log_2 \log_2 |\cP_{\delta}|)$ bits. 
\end{proof}

\subsection{Proof of Theorem~\ref{thm:info_theoretic_upper_bound}}
\begin{table}[H]
\centering
\begin{tabular}{|l|l|l|l|l|l|}
\hline
$\cH \downarrow, \cC \rightarrow$ & $c_1$ & $c_2$ & $c_3$ & \dots & $c_{|\cC|}$ \\ \hline
$h_1$                                                                                        & 1          & 0    & 0    &                       & 0                           \\\hline
$h_2$                                                                                        & 1          & 1    & 1    &                       & 1                           \\\hline
$h_3$                                                                                        & 0          & 1    & 0    &                       & 1                           \\\hline
\vdots                                                                       &            &      &      &                       &                             \\\hline
$h_{|\cH|}$                                                                 & 0          & 1    & 0    &                       & 1            \\\hline              
\end{tabular}%
\vspace{2em}
\caption{(Upper bound) Each entry of this table is i.i.d. Bernoulli with probability 0.5. All we need to do is to ensure that if we choose any $n$ columns above and any labeling in $\{0,1\}^n$, there exists one row that has that same labeling.}
\end{table}
\begin{proof}
Create a $\delta/2$ covering $\cC_{\delta/2}$ of the set $\cS$. We can partition the sphere into $|\cC_{\delta/2}|$ regions of diameter less than $\delta$: Any point that lies within distance $\delta/2$ to any point in $\cC_{\delta/2}$ is included in the region of that point. If a point lies within distance $\delta/2$ of multiple points in $\cC_{\delta/2}$, then assign it to any one of the regions arbitrarily. Since by assumption, no two points in our dataset are within distance $\delta$, the benefit of this partition is that every pair of points in our dataset will lie in different regions of the sphere. Consider the set of all the hypotheses $h_i$ that assign the same label to all the points which lie in a region of the partition. (If $\cS$ is a sphere, then these hypotheses look like a football with black and white regions:~black region is 0, white is 1). There will be $2^{\cC_{\delta/2}}$ such hypotheses. We will only select a random subset of this large set and denote it by $\cH$. We will show that $\log_2 \cH = \cO(n+ \log_2\log_2 |\cC_{\delta/2}|)$ will suffice for memorization.

For any particular choice of $n$ points and $n$ labels, the probability that none of the $|\cH|$ hypothesis have that choice of labels is $(1-2^{-n})^{|\cH|}$. Taking a union bound over all the $\binom{|\cC_{\delta/2}|}{n}$ choices of points and $2^n$ choices of labels, we need
\begin{align*}
\binom{|\cC_{\delta/2}|}{n} 2^n(1-2^{-n})^{|\cH|}<1,
\end{align*}
to ensure that there exists one $\cH$ for which we can memorize any $n$ points. Noting that $1-2^{-n}<e^{-2^{-n}}$ and $\binom{|\cC_{\delta/2}|}{n}<{|\cC_{\delta/2}|}^{n}$, we see that it is sufficient to have
\begin{align*}
(2|\cC_{\delta/2}|)^{n} e^{-2^{-n}{|\cH|}}<1.
\end{align*}
Solving this, we get that $\log_2 |\cH| >n+\log_2n + \log_2 \log_e 2|\cC_{\delta/2}|$ is sufficient, {\it i.e.}, we need $O(n+ \log_2\log_2 |\cC_{\delta/2}|)$ bits to memorize $n$ points.
\end{proof}

%% file: main.bbl
\begin{thebibliography}{25}
\providecommand{\natexlab}[1]{#1}
\providecommand{\url}[1]{\texttt{#1}}
\expandafter\ifx\csname urlstyle\endcsname\relax
  \providecommand{\doi}[1]{doi: #1}\else
  \providecommand{\doi}{doi: \begingroup \urlstyle{rm}\Url}\fi

\bibitem[Arpit et~al.(2017)Arpit, Jastrz{\k{e}}bski, Ballas, Krueger, Bengio,
  Kanwal, Maharaj, Fischer, Courville, Bengio, et~al.]{arpit2017closer}
Devansh Arpit, Stanis{\l}aw Jastrz{\k{e}}bski, Nicolas Ballas, David Krueger,
  Emmanuel Bengio, Maxinder~S Kanwal, Tegan Maharaj, Asja Fischer, Aaron
  Courville, Yoshua Bengio, et~al.
\newblock A closer look at memorization in deep networks.
\newblock In \emph{International Conference on Machine Learning}, pp.\
  233--242. PMLR, 2017.

\bibitem[Arratia \& Gordon(1989)Arratia and Gordon]{arratia1989tutorial}
Richard Arratia and Louis Gordon.
\newblock Tutorial on large deviations for the binomial distribution.
\newblock \emph{Bulletin of mathematical biology}, 51\penalty0 (1):\penalty0
  125--131, 1989.

\bibitem[Ball et~al.(1997)]{ball1997elementary}
Keith Ball et~al.
\newblock An elementary introduction to modern convex geometry.
\newblock \emph{Flavors of geometry}, 31:\penalty0 1--58, 1997.

\bibitem[Bartlett(1998)]{bartlett1998sample}
Peter~L Bartlett.
\newblock The sample complexity of pattern classification with neural networks:
  the size of the weights is more important than the size of the network.
\newblock \emph{IEEE transactions on Information Theory}, 44\penalty0
  (2):\penalty0 525--536, 1998.

\bibitem[Baum(1988)]{baum1988capabilities}
Eric~B Baum.
\newblock On the capabilities of multilayer perceptrons.
\newblock \emph{Journal of complexity}, 4\penalty0 (3):\penalty0 193--215,
  1988.

\bibitem[Belkin et~al.(2020)Belkin, Hsu, and Xu]{belkin2020two}
Mikhail Belkin, Daniel Hsu, and Ji~Xu.
\newblock Two models of double descent for weak features.
\newblock \emph{SIAM Journal on Mathematics of Data Science}, 2\penalty0
  (4):\penalty0 1167--1180, 2020.

\bibitem[Bubeck et~al.(2020)Bubeck, Eldan, Lee, and
  Mikulincer]{bubeck2020network}
S{\'e}bastien Bubeck, Ronen Eldan, Yin~Tat Lee, and Dan Mikulincer.
\newblock Network size and weights size for memorization with two-layers neural
  networks.
\newblock \emph{arXiv preprint arXiv:2006.02855}, 2020.

\bibitem[Cover(1965)]{cover1965geometrical}
Thomas~M Cover.
\newblock Geometrical and statistical properties of systems of linear
  inequalities with applications in pattern recognition.
\newblock \emph{IEEE transactions on electronic computers}, 1965.

\bibitem[Guruswami et~al.(2012)Guruswami, Rudra, and
  Sudan]{guruswami2012essential}
Venkatesan Guruswami, Atri Rudra, and Madhu Sudan.
\newblock Essential coding theory.
\newblock \emph{Draft available at http://www. cse. buffalo. edu/~
  atri/courses/coding-theory/book}, 2012.

\bibitem[Hardt \& Ma(2017)Hardt and Ma]{hardt2017identity}
Moritz Hardt and Tengyu Ma.
\newblock Identity matters in deep learning.
\newblock In \emph{5th International Conference on Learning Representations,
  {ICLR} 2017, Toulon, France, April 24-26, 2017, Conference Track
  Proceedings}, 2017.

\bibitem[Huang(2003)]{huang2003learning}
Guang-Bin Huang.
\newblock Learning capability and storage capacity of two-hidden-layer
  feedforward networks.
\newblock \emph{IEEE transactions on neural networks}, 14\penalty0
  (2):\penalty0 274--281, 2003.

\bibitem[Huang et~al.(1991)Huang, Huang, et~al.]{huang1991bounds}
Shih-Chi Huang, Yih-Fang Huang, et~al.
\newblock Bounds on the number of hidden neurons in multilayer perceptrons.
\newblock \emph{IEEE transactions on neural networks}, 2\penalty0 (1):\penalty0
  47--55, 1991.

\bibitem[Kotsovsky et~al.(2020)Kotsovsky, Geche, and
  Batyuk]{kotsovsky2020bithreshold}
Vladyslav Kotsovsky, Fedir Geche, and Anatoliy Batyuk.
\newblock Bithreshold neural network classifier.
\newblock In \emph{2020 IEEE 15th International Conference on Computer Sciences
  and Information Technologies (CSIT)}, volume~1, pp.\  32--35. IEEE, 2020.

\bibitem[Kowalczyk(1997)]{kowalczyk1997estimates}
Adam Kowalczyk.
\newblock Estimates of storage capacity of multilayer perceptron with threshold
  logic hidden units.
\newblock \emph{Neural networks}, 10\penalty0 (8):\penalty0 1417--1433, 1997.

\bibitem[Liu et~al.(2020)Liu, Papailiopoulos, and Achlioptas]{liu2020bad}
Shengchao Liu, Dimitris Papailiopoulos, and Dimitris Achlioptas.
\newblock Bad global minima exist and sgd can reach them.
\newblock \emph{Advances in Neural Information Processing Systems}, 33, 2020.

\bibitem[Mitchison \& Durbin(1989)Mitchison and Durbin]{mitchison1989bounds}
GJ~Mitchison and RM~Durbin.
\newblock Bounds on the learning capacity of some multi-layer networks.
\newblock \emph{Biological Cybernetics}, 1989.

\bibitem[Neyshabur et~al.(2018)Neyshabur, Li, Bhojanapalli, LeCun, and
  Srebro]{neyshabur2018role}
Behnam Neyshabur, Zhiyuan Li, Srinadh Bhojanapalli, Yann LeCun, and Nathan
  Srebro.
\newblock The role of over-parametrization in generalization of neural
  networks.
\newblock In \emph{International Conference on Learning Representations}, 2018.

\bibitem[Neyshabur et~al.(2019)Neyshabur, Li, Bhojanapalli, LeCun, and
  Srebro]{neyshabur2019towards}
Behnam Neyshabur, Zhiyuan Li, Srinadh Bhojanapalli, Yann LeCun, and Nathan
  Srebro.
\newblock Towards understanding the role of over-parametrization in
  generalization of neural networks.
\newblock In \emph{International Conference on Learning Representations
  (ICLR)}, 2019.

\bibitem[Park et~al.(2020)Park, Lee, Yun, and Shin]{park2020provable}
Sejun Park, Jaeho Lee, Chulhee Yun, and Jinwoo Shin.
\newblock Provable memorization via deep neural networks using sub-linear
  parameters.
\newblock \emph{arXiv preprint arXiv:2010.13363}, 2020.

\bibitem[Sartori \& Antsaklis(1991)Sartori and Antsaklis]{sartori1991simple}
Michael~A Sartori and Panos~J Antsaklis.
\newblock A simple method to derive bounds on the size and to train multilayer
  neural networks.
\newblock \emph{IEEE transactions on neural networks}, 2\penalty0 (4):\penalty0
  467--471, 1991.

\bibitem[Sontag(1990)]{sontag1990remarks}
Eduardo~D Sontag.
\newblock Remarks on interpolation and recognition using neural nets.
\newblock In \emph{NIPS}, pp.\  939--945, 1990.

\bibitem[Vershynin(2018)]{vershynin2018high}
Roman Vershynin.
\newblock \emph{High-dimensional probability: An introduction with applications
  in data science}, volume~47.
\newblock Cambridge university press, 2018.

\bibitem[Vershynin(2020)]{vershynin2020memory}
Roman Vershynin.
\newblock Memory capacity of neural networks with threshold and rectified
  linear unit activations.
\newblock \emph{SIAM Journal on Mathematics of Data Science}, 2\penalty0
  (4):\penalty0 1004--1033, 2020.

\bibitem[Yun et~al.(2019)Yun, Sra, and Jadbabaie]{yun2018small}
Chulhee Yun, Suvrit Sra, and Ali Jadbabaie.
\newblock Small relu networks are powerful memorizers: a tight analysis of
  memorization capacity.
\newblock In \emph{Advances in Neural Information Processing Systems 32: Annual
  Conference on Neural Information Processing Systems 2019, NeurIPS 2019,
  December 8-14, 2019, Vancouver, BC, Canada}, pp.\  15532--15543, 2019.

\bibitem[Zhang et~al.(2017)Zhang, Bengio, Hardt, Recht, and
  Vinyals]{Zhang17understanding}
Chiyuan Zhang, Samy Bengio, Moritz Hardt, Benjamin Recht, and Oriol Vinyals.
\newblock Understanding deep learning requires rethinking generalization.
\newblock In \emph{5th International Conference on Learning Representations,
  {ICLR} 2017, Toulon, France, April 24-26, 2017, Conference Track
  Proceedings}, 2017.

\end{thebibliography}
